\documentclass[review]{elsarticle}

\usepackage{lineno,hyperref}
\usepackage{amssymb,amsmath,amsthm,latexsym,amscd,amsfonts}
\usepackage{dsfont}
\usepackage{soul}
\usepackage{caption}
\usepackage{algorithm}
\usepackage{algpseudocode}

\usepackage{tikz}
\usetikzlibrary{%
  arrows,%
  shapes.misc,
  shapes.arrows,%
  chains,%
  matrix,%
  positioning,
  scopes,%
  decorations.pathmorphing,
  shapes.geometric,
  shadows,
  decorations.pathreplacing,
  patterns,
  shapes,
  decorations.markings
}
\usepackage{tabularx}
\usepackage{graphicx}
\graphicspath{ {./} }
\usepackage{epstopdf}
\usepackage{xcolor,colortbl}

\modulolinenumbers[5]

\begin{document}

\begin{frontmatter}

\title{Adjustable Privacy using Autoencoder-based Learning Structure}

\author{Mohammad A. Jamshidi$^\star$}
\author{Hadi Veisi$^\dagger$}
\author{Mohammad M. Mojahedian$^\star$}
\author{Mohammad R. Aref$^\star$}
\address{$^\star$Information Systems and Security Lab. (ISSL), Sharif University of Tech., Tehran, Iran\\$^\dagger$Faculty of New Sciences and Technologies, University of Tehran, Tehran, Iran}
\address{m.a.jamshidi992@gmail.com, h.veisi@ut.ac.ir, m.mojahedian@gmail.com, aref@sharif.ir}

%
%

\begin{abstract}
Inference centers need more data to have a more comprehensive and beneficial learning model, and for this purpose, they need to collect data from data providers. On the other hand, data providers are cautious about delivering their datasets to inference centers in terms of privacy considerations. In this paper, by modifying the structure of the autoencoder, we present a method that manages the utility-privacy trade-off well. To be more precise, the data is first compressed using the encoder, then confidential and non-confidential features are separated and uncorrelated using the classifier. The confidential feature is appropriately combined with noise, and the non-confidential feature is enhanced, and at the end, data with the original data format is produced by the decoder. The proposed architecture also allows data providers to set the level of privacy required for confidential features. The proposed method has been examined for both image and categorical databases, and the results show a significant performance improvement compared to previous methods.
\end{abstract}


\end{frontmatter}


\section{Introduction}
\label{sec:introduction}
\begin{figure}[h]
\centering
\begin{tikzpicture}[scale=0.8]

\draw[thick] (0,2) rectangle (1,3) node [pos=.5] {$\text{DP}_1$};

\draw[thick] (0,0) rectangle (1,1) node [pos=.5] {$\text{DP}_2$};

\draw[thick,dotted] (.5,-1.5) -- (.5,-.5);

\draw[thick] (0,-3) rectangle (1,-2) node [pos=.5] {$\text{DP}_m$};

\draw[thick] (3,-.65) rectangle (6.1,.65) node [align=center,pos=.5] {Utility\\[-1ex]Provider};

\draw[-latex] (1,.5) -- (3,.5);
\draw[dashed,-latex] (3,.4) -- (1,.4);

\draw (1,2.5) -- (3.75,2.5);
\draw[-latex] (3.75,2.5) -- (3.75,.65);
\draw[dashed,latex-] (1,2.4) -- (3.65,2.4);
\draw[dashed] (3.65,2.4) -- (3.65,.65);

\draw (1,-2.5) -- (3.75,-2.5);
\draw[-latex] (3.75,-2.5) -- (3.75,-.65);
\draw[dashed,latex-] (1,-2.4) -- (3.75,-2.4);
\draw[dashed] (3.65,-2.4) -- (3.65,-.65);

\draw[thick] (7,-.5) rectangle (8.5,.5) node [pos=.5] {Users};

\draw[thick,dashed,-latex] (6.1,0) -- (7,0);

\end{tikzpicture}
\caption{Data providers (DPs) tend to distort their data and send it to the utility provider in a way that balances privacy and utility.}
\label{fig:data_privacy}
\end{figure}
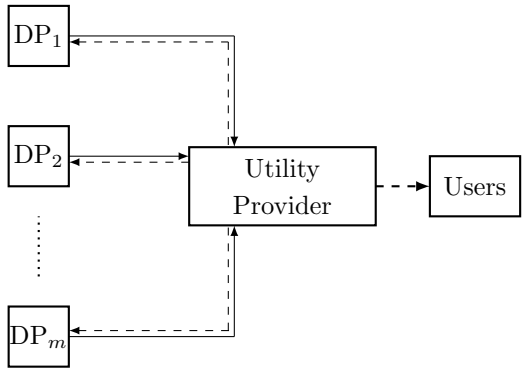

The more data a learning system can access, its model can be more comprehensive. But sometimes, the learner or utility provider does not have access to much data or does not have any data at all and must receive it from different units. Different units may be sensitive about their data and do not like to provide all their information to the utility provider. In this paper, we are looking for a solution so that data providers can distort their data to such an extent that the utility provider can use it and that the information they want remains confidential as much as possible from the utility provider or any other adversary. To be more precise, as shown in Fig. \ref{fig:data_privacy}, we have $m$ data providers, each of which has a set of data and wants to provide it to the utility provider in such a way that a part of the data or its features can remain private from the utility provider or any other adversary. On the other hand, the quality of the data provided to the utility provider must be good enough to train his learning model well. The utility provider is supposed to provide services using the resulting model to data providers or other users. Therefore the trade-off between the utility and privacy of the datasets becomes important.

Ensuring privacy is paramount in an era where so much data is available. The utility-privacy trade-off has been studied theoretically \cite{sankar2013utility,kalantari2017information} and algorithmically. Various algorithms have been proposed to balance the trade-off, and the present paper is in this direction. The primary algorithms proposed for privacy include $k$-anonymity \cite{sweeney2002k}, $\ell$-diversity \cite{machanavajjhala2007diversity}, and $t$-clossness \cite{li2006t}, which are suitable only for small datasets. Another group of private algorithms is based on differential privacy, which is a mathematical tool that guarantees the privacy of the dataset by appropriately adding noise \cite{dwork2014algorithmic,alvim2011differential,abadi2016deep,phan2016differential,papernot2016semi}. Among the difficulties of differential privacy for high dimensions, it is time-consuming, requires a lot of noise and as a result distorts the dataset utility, etc. There are other private methods, including homomorphic encryption \cite{hesamifard2017cryptodl,emekcci2007privacy}, and secure multi-party computation (SMPC) \cite{mohassel2017secureml,tran2021efficient}, which are two cryptographic approaches that are related to calculations on encrypted data and face limitations such as computational cost, limited range of computational operations that can be performed, communication cost, etc. In addition to the centralized privacy-preserving methods, there is the distributed federated learning framework, where each data provider trains the model. This method faces limitations such as the negative effect of heterogeneous systems of different data providers, inefficiency in terms of communication, etc. \cite{shokri2015privacy,zhu2021distributed,wei2020federated}.

Another category of private algorithms that have received special attention with the increase in data dimensions are heuristic methods based on machine learning tools \cite{kuang2021unnoticeable,raval2019olympus,
huang2017context,kairouz2022generating,
li2020tiprdc,li2019deepobfuscator,
singh2022decouple,osia2018deep,osia2020hybrid,
nguyen2020autogan,mandal2022uncertainty,
louizos2015variational,song2019learning,chen2019distributed}. These methods use machine learning tools such as autoencoders, generative adversarial networks (GAN), variational autoencoders, etc. This paper is also included in this category. In particular, we discuss privacy in data publishing using neural networks. There are similar works in this direction. Huang et al. in \cite{huang2017context,kairouz2022generating} uses the generative adversarial network (GAN) and solves a minimax game between the privatizer and the adversary to conceal a \underline{specific} feature. Li et al. in \cite{li2020tiprdc} have used mutual information-based training to learn the feature extractor so that the private features are hidden while maintaining the utility of the remaining information. The output of the proposed method in \cite{li2020tiprdc} is a censored feature vector that is not in the original data format and, therefore, not suitable for data publishing, which makes it useless for pre-trained models such as DenseNet. In \cite{singh2022decouple}, by combining variational autoencoder and differential privacy, the data dimension is first reduced, then sensitive and non-sensitive data are separated using two classifiers, and after the covariance matrix of sensitive data is learned, it is perturbed to establish differential privacy. Using the differential privacy tool requires more training data and is therefore time-consuming \cite{tramer2020differentially}. A feature extraction method in \cite{osia2020hybrid} is proposed for implementation on mobile devices based on Siamese architecture. This method also produces data that does not match the original dataset format. In \cite{nguyen2020autogan}, the obfuscator uses an autoencoder to reduce the dimensions of the image, and then with a GAN-based structure, the encoder output distribution approaches the Gaussian distribution. Also, the classifier that extracts the desired feature gives feedback to the obfuscator. Mandal et al. proposed a private learning algorithm based on uncertainty autoencoder in \cite{mandal2022uncertainty}. Their work has significant results on categorical datasets, so we use it as a benchmark for comparison. Making the dataset confidential may lead to a dataset that does not have a standard format; for this purpose, in \cite{mandal2022uncertainty}, the algorithm's performance with data-type ignorant and data-type aware conditions is studied.

In this paper, we use the autoencoder-based structure, which helps us in two ways by reducing the data dimension. First, the encoder output will be compressed data that are as uncorrelated as possible. Secondly, processing can be done on the reduced-dimension data with a simpler network, significantly reducing the computational load. The paper's main idea is that the dimensionality-reduced data is appropriately processed using neural networks so that while the utility of the dataset is maintained in terms of some features, the obfuscated dataset is private in terms of other features. The advantages and contributions of the proposed private learning algorithm are:
\begin{itemize}
\item The proposed scheme works well on both categorical and image datasets.
\item In terms of utility-privacy trade-off, the proposed algorithm outperforms other methods while using a much simpler structure that makes it more suitable for use on weaker processors such as mobile and Internet of Things (IoT) devices.

\item The obfuscating model can be learned by a utility provider or any trusted entity and then sent to the data providers for use. This removes the burden of training the model from the data providers. It is also possible for each data provider to adjust the level of privacy they need without changing the model and just by tuning the noise. Further, the data providers are given a parameter to adjust the data utility amount.

\item In our proposed obfuscator, all the features are obscured except for non-confidential ones, which makes it more private against adversaries who do not have a specific goal. To put it more clearly, in our method, the feature or features that the utility provider wants to infer are \underline{known}, and the data providers obscure the rest of the features.

\end{itemize}

The rest of the paper is organized as follows. The system model is described in Section \ref{sec:system_model}. The details of the proposed structure are discussed in Section \ref{sec:methodology}. Simulation results are included in Section \ref{sec:simulation}. Finally, the paper is concluded in Section \ref{sec:conclusion}.

\section{System Model}
\label{sec:system_model}
The dataset $\mathcal{D}$ is a collection containing $n$ samples of the instance space $\mathcal{Z}=\mathcal{X}\times\mathcal{Y}$, in which $\mathcal{X}$ and $\mathcal{Y}$ are input and output spaces, respectively. We assume that $\mathbf{Y}\in\mathcal{Y}$ is a vector of features that are divided into private and non-private categories in the form of $\mathbf{Y}=\left(\mathbf{Y}_\mathsf{P},\mathbf{Y}_\mathsf{NP}\right)$. For example, in this paper, we consider a set of face images as $\mathcal{X}$, the smiling feature as non-private and other features as private.

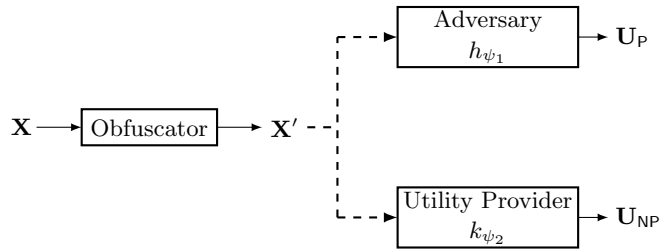
\begin{figure}[h]
\centering
\begin{tikzpicture}[scale=0.8]

\tikzstyle{every node}=[font=\small]

\draw[thick] (1.75,-.3) rectangle (4,.3) node [pos=.5] {Obfuscator};

\draw[-latex] (1,0) -- (1.75,0) node at (0.75,0) {$\mathbf{X}$};
\draw[-latex] (4,0) -- (4.75,0) node at (5.125,0) {$\mathbf{X}'$};

\draw[thick,dashed] (5.5,0) -- (6,0);
\draw[thick,dashed] (6,-1.5) -- (6,1.5);

\draw[thick,dashed,-latex] (6,-1.5) -- (7,-1.5);
\draw[thick] (7,-2) rectangle (10,-1) node [pos=.5,align=center] {Utility Provider\\[-1.3ex]$k_{\psi_2}$};

\draw[thick,dashed,-latex] (6,1.5) -- (7,1.5);
\draw[thick] (7,2) rectangle (10,1) node [pos=.5,align=center] {Adversary\\[-1.3ex]$h_{\psi_1}$};

\draw[-latex] (10,1.5) -- (10.5,1.5) node at (10.9,1.5) {$\mathbf{U}_\mathsf{P}$};
\draw[-latex] (10,-1.5) -- (10.5,-1.5) node at (11,-1.5) {$\mathbf{U}_\mathsf{NP}$};

\end{tikzpicture}
\caption{Adversary and utility provider are two deep neural networks with $\psi_1$ and $\psi_2$ parameters, respectively, which receive obfuscated data as input and try to extract private and non-private features, respectively.}
\label{fig:system_model_AU}
\end{figure}

The data provider intends to deliver data to the utility provider for collaborative learning, and on the other hand, the confidentiality of some features is important to him. Therefore, by converting $\mathbf{X}$ to $\mathbf{X}'$, the data provider aims to keep all features private except the non-private one while the utility of the dataset is acceptable in terms of the non-private feature. The function that converts $\mathbf{X}$ to $\mathbf{X}'$ is an obfuscator; the amount of ambiguity it adds to the dataset determines the utility-privacy trade-off. The Obfuscator, adversary, and utility provider are shown in Fig. \ref{fig:system_model_AU}. Here, $\mathbf{U}_\mathsf{P}$ and $\mathbf{U}_\mathsf{NP}$ are private and non-private features, respectively, that are inferred by the adversary and the utility provider from the obfuscated dataset.

\underline{Threat Model}\\
We assume the adversary is a machine learning model that seeks to extract a private feature from the obfuscated dataset. We also assume that the data provider does not know what feature or features the adversary looks for. Therefore, the data provider tries to obfuscate all features other than the utility provider's demands. In this paper, we consider two types of adversaries.
\begin{itemize}
\item Weak adversary: This type of adversary does not have access to the obfuscation model and only has the obfuscated dataset. Therefore, it is trained using a dataset similar to the original dataset (non-obfuscated) and then extracts the desired feature from the obfuscated dataset.
\item Strong adversary: This type of adversary introduced in \cite{raval2017protecting} has access to obfuscator model. Therefore, it creates an obfuscated dataset from a dataset similar to the original dataset. Now it has a dummy obfuscated dataset, and on the other hand, it has the exact value of the feature it wants to infer and trains its network using supervised learning.
\end{itemize}

\underline{Utility Provider}\\
A utility provider is a deep neural network (DNN) classifier trained on the obfuscated dataset to infer one or more features. Other users can utilize this network to infer features from the unobfuscated dataset available to them. It is assumed that the data provider is \underline{aware} of non-private features.

\begin{figure}[h]
\centering
\begin{tikzpicture}[scale=.75]

\draw[-latex] (1.5,0) -- (2.25,0) node at (1.2,0) {$\mathbf{X}$};
\draw[thick] (2.25,-.5) rectangle (3.25,.5) node [pos=.5] {$\mathsf{E}_\alpha$};
\draw[thick,dashed] (3.25,0) -- (4.75,0) node at (4,0.25) {$\mathbf{V}$};

\draw[thick,dashed] (4.75,-1.5) -- (4.75,1.5);

\draw[thick,dashed,-latex] (4.75,-1.5) -- (5.75,-1.5);
\draw[thick] (5.75,-2) rectangle (6.75,-1) node [pos=.5,align=center] {$\mathsf{C}_\theta$};

\draw[thick,dashed,-latex] (4.75,1.5) -- (5.75,1.5);
\draw[thick] (5.75,1) rectangle (6.75,2) node [pos=.5,align=center] {$\mathsf{R}_\phi$};

\draw[-latex] (6.75,1.5) -- (8.25,1.5) node at (7.5,1.75) {$\mathbf{W}_\mathsf{P}$};
\draw[-latex] (6.75,-1.5) -- (8.25,-1.5) node at (7.5,-1.25) {$\mathbf{W}_\mathsf{NP}$};

\draw[thick] (8.25,1) rectangle (9.25,2) node [pos=.5,align=center] {$f(\cdot)$};

\draw[thick] (8.25,-2) rectangle (9.25,-1) node [pos=.5,align=center] {$g(\cdot)$};

\draw[thick] (9.25,1.5) -- (10.75,1.5) node at (10,1.8) {$\mathbf{W}'_\mathsf{P}$};
\draw[thick] (9.25,-1.5) -- (10.75,-1.5) node at (10,-1.2) {$\mathbf{W}'_\mathsf{NP}$};

\draw[thick] (10.75,0.25) -- (10.75,1.5);
\draw[thick] (10.75,-0.25) -- (10.75,-1.5);

\draw[thick,-latex] (10.75,-0.25) -- (11.4,-0.25);
\draw[thick,-latex] (10.75,0.25) -- (11.4,0.25);

\draw[thick] (11.4,-0.5) rectangle (12.4,0.5) node [pos=.5,align=center] {$\parallel$};

\draw[thick,-latex] (12.4,0) -- (13.9,0) node at (13.15,0.25) {$\mathbf{V}'$};


\draw[thick] (13.9,-0.5) rectangle (14.9,0.5) node [pos=.5,align=center] {$\mathsf{D}_\beta$};

\draw[thick,-latex] (14.9,0) -- (15.9,0) node at (16.2,0) {$\mathbf{X}'$};

\end{tikzpicture}
\caption{The obfuscator consists of an autoencoder, inside which the private and non-private features are intelligently separated. Then, the private features are combined with the noise appropriately while the non-private features are enhanced. Then the decoder constructs the obfuscated dataset from all the manipulated features.}
\label{fig:obf_model}
\end{figure}
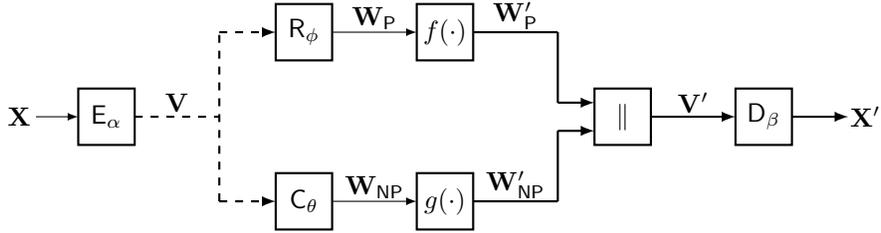

\underline{Obfuscator}\\
The paper's main idea is to decorrelate private and non-private features and then obfuscate the private features while enhancing the non-private ones. For this purpose, as shown in Fig. \ref{fig:obf_model}, the dimension of the data is reduced by using the autoencoder, and decorrelation is done inside it. We employ autoencoder for three reasons:
\begin{enumerate}
\item By reducing the dimension, the autoencoder reduces the utility of the data and increases its privacy.
\item Encoder produces uncorrelated output by compressing data and removing redundancy.
\item By reducing the dimension of the data, the intermediate networks that have the task of separating non-private features from compressed data will have a much simpler structure.
\end{enumerate}

The autoencoder consists of encoder $\mathsf{E}_\alpha$ and decoder $\mathsf{D}_\beta$, and the coded data is determined by $\mathbf{V}$. More precisely,
\begin{align}
\mathbf{V}=\mathsf{E}_\alpha\!\left(\mathbf{X}\right).
\end{align}
Here, the encoder and decoder are DNNs whose parameter sets are indicated by $\alpha$ and $\beta$, respectively.

Now, we make two sets of data from the reduced dimensional and meaningless coded data $\mathbf{V}$.
\begin{enumerate}
\item The DNN $\mathsf{C}_\theta$ with parameter $\theta$: It is actually a classifier that takes meaningless data $\mathbf{V}$ and produces meaningful features $\mathbf{W}_\mathsf{NP}=\mathsf{C}_\theta\left(\mathbf{V}\right)$. The features represented by $\mathbf{W}_\mathsf{NP}$ are the ones we want to remain useful.
\item The DNN $\mathsf{R}_\phi$ with parameter $\phi$: Its output contains information from $\mathbf{V}$ that is appropriately uncorrelated from $\mathbf{W}_\mathsf{NP}$. We want to keep this part of the information as confidential as possible, so we have shown it with $\mathbf{W}_\mathsf{P}$.
\end{enumerate}
In the next step, we apply the functions $f(\cdot)$ and $g(\cdot)$ to obtain the data $\mathbf{W}'_\mathsf{P}$ and $\mathbf{W}'_\mathsf{NP}$, respectively, and create $\mathbf{V}'$ by putting them together. In the final step, the decoder will convert $\mathbf{V}'$ to $\mathbf{X}'$. Now we have to design parameters $\alpha$, $\beta$, $\phi$, $\theta$ and functions $f(\cdot)$ and $g(\cdot)$. For this purpose, we consider the following properties to balance privacy and utility.
\begin{itemize}
\item[(P1)] To preserve the utility of the original dataset, we like $\mathbf{X}$ and $\mathbf{X}'$ to be as similar as possible. For this purpose, we will minimize the following loss function.
\begin{align}
L_{ae}(\alpha,\beta,\phi,\theta)= \mathbb{E}\Big[\ell(\mathbf{X},\mathbf{X'})\Big],
\end{align}
where the expectation is over the distribution of $\mathbf{X}$. Additionally, $\mathbf{X}'$ without considering the functions $f(\cdot)$ and $g(\cdot)$ is equal to
\begin{align}
\mathbf{X'} = \mathrm{D}_{\beta}\big(\mathrm{C}_{\theta}(\mathbf{V})\parallel\mathrm{R}_{\phi}(\mathbf{V})\big).
\end{align}

\item[(P2)] To make $\mathbf{W}_\mathsf{NP}$ useful for utility provider, we consider minimizing the following.
\begin{align}
L_{\mathsf{C}}(\alpha,\theta)= \mathbb{E}\Big[\ell\left(\mathbf{W}_\mathsf{NP},\mathbf{Y}_\mathsf{NP}\right)\Big],
\end{align}
where the expectation is over the joint distribution of $\mathbf{X}$ and $\mathbf{Y}$.


\item[(P3)] Suppose that adversary $h_{\psi_1}$ is a DNN with parameter $\psi_1$ that is well trained on a dataset similar to the original dataset/dummy obfuscated dataset (weak/strong) to infer private features $\mathbf{Y}_\mathsf{P}$. As a result, the adversary wants to minimize the following loss function, while we want it to be maximized.
\begin{align}
L_{h_{\psi_1}}(f,g)= \mathbb{E}\left[\ell\Big(h_{\psi_1}\left(\mathsf{D}_\beta\left(\mathbf{W}'_\mathsf{NP}\parallel \mathbf{W}'_\mathsf{P}\right)\right),\mathbf{Y}_\mathsf{P}\Big)\right].
\end{align}

\item[(P4)] Suppose that utility provider $k_{\psi_2}$ is a DNN with parameter $\psi_2$ that is well trained on the obfuscated dataset to infer certain non-private feature $\mathbf{Y}_\mathsf{NP}$. As a result, the utility provider wants to minimize the following loss function.
\begin{align}
L_{k_{\psi_2}}(f,g)= \mathbb{E}\left[\ell\Big(k_{\psi_2}\left(\mathsf{D}_\beta\left(\mathbf{W}'_\mathsf{NP}\parallel \mathbf{W}'_\mathsf{P}\right)\right),\mathbf{Y}_\mathsf{NP}\Big)\right].
\end{align}
\end{itemize}

The design of the proposed network is done in two stages. In the first step, assuming the absence of functions $f(\cdot)$ and $g(\cdot)$ and using (P1) and (P2), the network parameters are calculated as follows.
\begin{align}
\label{eqn:DNN_train_Loss}
\alpha^*,\beta^*,\phi^*,\theta^* = \arg \min_{\alpha,\beta,\phi,\theta}\,L_{ae}(\alpha,\beta,\phi,\theta)+L_{\mathsf{C}}(\alpha,\theta). 
\end{align}
Then $f(\cdot)$ and $g(\cdot)$ are optimized according to the following optimization problem.
\begin{align}
\label{eqn:DNN_func_Loss}
(f^*,g^*) = \arg \max_{f,g}\,L_{h_{\psi_1}}(f,g)-L_{k_{\psi_2}}(f,g).
\end{align}

\section{Methodology}
\label{sec:methodology}

As mentioned in the previous section, \eqref{eqn:DNN_train_Loss} is used to train DNNs $\mathsf{E}_\alpha$, $\mathsf{D}_\beta$, $\mathsf{C}_\theta$ and $\mathsf{R}_\phi$. In addition, instead of solving the optimization problem in \eqref{eqn:DNN_func_Loss}, we design the functions $f(\cdot)$ and $g(\cdot)$ intelligently. Further, the proposed scheme is examined on two image and categorical datasets to measure its performance in various applications.

\subsection{Image Dataset}

CelebA is chosen as the dataset, a collection of large-scale facial features \cite{liu2015deep}. Facial images with size $64\times64\times3$ are input, and their labels are different features, including age, gender, etc. To compare the performance of the proposed scheme with the previous works, the desired features of open mouth, smiling, and high cheekbone have been selected, which are denoted by ``CelebA-G-M", ``CelebA-G-S", ``CelebA-G-C", respectively. Here ``G" stands for gender, which is the feature the adversary is looking for. It should be noted that we assume that the data providers do not know which feature is confidential.

\underline{DNNs Structures}\\
Inspired by VGG-$16$ network \cite{simonyan2014very}, the building blocks of the autoencoder, i.e. $\mathsf{E}_\alpha$ and $\mathsf{D}_\beta$, both consist of $4$ $2$D-convolutional layers, $3$ batch normalization layers, and $1$ fully-connected layer. $\mathsf{E}_\alpha$ takes images with size $64\times64\times3$ and produces $1024$ features as output. In other words, the $\mathbf{V}$ size equals $1024$. The decoder then converts a vector of manipulated features with size $1024$ into a $64\times64\times3$ image.

Two fully-connected networks $\mathsf{C}_\theta$ and $\mathsf{R}_\phi$ are used in the middle of the autoencoder. $\mathsf{C}_\theta$ is a $4$-layer fully-connected network that converts the input of size $1024$ into $2$ outputs expressing the desired feature of the utility provider. $\mathsf{R}_\phi$ is a $3$-layer fully-connected network that converts the input of size $1024$ into $1022$ outputs representing the rest of the features that are well uncorrelated with the desirable feature of the utility provider. The DNNs structure details are given in Table \ref{tab:1}.
\begin{table}[t!]
\centering
\caption{DNNs architecture details for image datasets.}
\resizebox{\columnwidth}{!}{
\begin{tabular}{cccccc}
\hline 
\rule[-1ex]{0pt}{2.5ex} Component & Num & Layer & Output Size & Specs & Activation Function \\ 
\hline 
\rule[-1ex]{0pt}{2.5ex} Input Data &   & Image Samples & $3\times64\times64$ &   \\ 
\hline 
\rule[-1ex]{0pt}{2.5ex} Encoder & 1 & Conv2D & $64\times32\times32$ & kernel=4, stride=2, padding=1 & LeakyReLU \\ 
\rule[-1ex]{0pt}{2.5ex}   & 2 & Conv2D & $64\times16\times16$ & kernel=4, stride=2, padding=1 &   \\ 
\rule[-1ex]{0pt}{2.5ex}   & 3 & BatchNorm2D &   & eps=1e-5, momentum=0.1 & LeakyReLU \\ 
\rule[-1ex]{0pt}{2.5ex}   & 4 & Conv2D & $64\times8\times8$ & kernel=4, stride=2, padding=1 &   \\ 
\rule[-1ex]{0pt}{2.5ex}   & 5 & BatchNorm2D &   & eps=1e-5, momentum=0.1 & LeakyReLU \\ 
\rule[-1ex]{0pt}{2.5ex}   & 6 & Conv2D & $128\times4\times4$ & kernel=4, stride=2, padding=1 &   \\ 
\rule[-1ex]{0pt}{2.5ex}   & 7 & BatchNorm2D & $128\times4\times4$ $\rightarrow$ 2048 (shaped) & eps=1e-5, momentum=0.1 & LeakyReLU \\ 
\rule[-1ex]{0pt}{2.5ex}   & 8 & Linear & 1024 &   & LeakyReLU \\ 
\hline 
\rule[-1ex]{0pt}{2.5ex} Classifier & 1 & Linear & 1024 & Dropout(p=0.5) & LeakyReLU \\ 
\rule[-1ex]{0pt}{2.5ex}   & 2 & Linear & 256 & Dropout(p=0.5) & LeakyReLU \\ 
\rule[-1ex]{0pt}{2.5ex}   & 3 & Linear & 64 &   & LeakyReLU \\ 
\rule[-1ex]{0pt}{2.5ex}   & 4 & Linear & 2 &   & LogSoftMax \\ 
\hline 
\rule[-1ex]{0pt}{2.5ex} R & 1 & Linear & 1024 & Dropout(p=0.5) & LeakyReLU \\ 
\rule[-1ex]{0pt}{2.5ex}   & 2 & Linear & 1024 & Dropout(p=0.5) & LeakyReLU \\ 
\rule[-1ex]{0pt}{2.5ex}   & 3 & Linear & 1022 &   & LeakyReLU \\ 
\hline 
\rule[-1ex]{0pt}{2.5ex} Decoder & 1 & Linear & 2048 $\rightarrow$ $128\times4\times4$ (shaped) &   & LeakyReLU \\ 
\rule[-1ex]{0pt}{2.5ex}   & 2 & ConvTranspose2D & $64\times8\times8$ & kernel=4, stride=2, padding=1 &   \\ 
\rule[-1ex]{0pt}{2.5ex}   & 3 & BatchNorm2D &   & eps=1e-5, momentum=0.1 & ReLU \\ 
\rule[-1ex]{0pt}{2.5ex}   & 4 & ConvTranspose2D & $64\times16\times16$ & kernel=4, stride=2, padding=1 &   \\ 
\rule[-1ex]{0pt}{2.5ex}   & 5 & BatchNorm2D &   & eps=1e-5, momentum=0.1 & ReLU \\ 
\rule[-1ex]{0pt}{2.5ex}   & 6 & ConvTranspose2D & $64\times32\times32$ & kernel=4, stride=2, padding=1 &   \\ 
\rule[-1ex]{0pt}{2.5ex}   & 7 & BatchNorm2D &   & eps=1e-5, momentum=0.1 & ReLU \\ 
\rule[-1ex]{0pt}{2.5ex}   & 8 & ConvTranspose2D & $3\times64\times64$ & kernel=4, stride=2, padding=1 & Sigmoid \\ 
\hline 
\rule[-1ex]{0pt}{2.5ex} Adversary & 1 & Conv2D & $64\times32\times32$ & kernel=4, stride=2, padding=1 & LeakyReLU \\ 
\rule[-1ex]{0pt}{2.5ex} and & 2 & Conv2D & $64\times16\times16$ & kernel=4, stride=2, padding=1 &   \\ 
\rule[-1ex]{0pt}{2.5ex} Utility Provider & 3 & BatchNorm2D &   & eps=1e-5, momentum=0.1 & LeakyReLU \\ 
\rule[-1ex]{0pt}{2.5ex}   & 4 & Conv2D & $64\times8\times8$ & kernel=4, stride=2, padding=1 &   \\ 
\rule[-1ex]{0pt}{2.5ex}   & 5 & BatchNorm2D &   & eps=1e-5, momentum=0.1 & LeakyReLU \\ 
\rule[-1ex]{0pt}{2.5ex}   & 6 & Conv2D & $128\times4\times4$ & kernel=4, stride=2, padding=1 &   \\ 
\rule[-1ex]{0pt}{2.5ex}   & 7 & BatchNorm2D & $128\times4\times4$ $\rightarrow$ 2048 (shaped) & eps=1e-5, momentum=0.1 & LeakyReLU \\ 
\rule[-1ex]{0pt}{2.5ex}   & 8 & Linear & 1024 &   & LeakyReLU \\ 
\rule[-1ex]{0pt}{2.5ex}   & 9 & Linear & 1024 & Dropout(p=0.5) & LeakyReLU \\ 
\rule[-1ex]{0pt}{2.5ex}   & 10 & Linear & 256 & Dropout(p=0.5) & LeakyReLU \\ 
\rule[-1ex]{0pt}{2.5ex}   & 11 & Linear & 64 &   & LeakyReLU \\ 
\rule[-1ex]{0pt}{2.5ex}   & 12 & Linear & 2 &   & LogSoftMax \\ 
\hline 
\end{tabular} 
}
\label{tab:1}
\end{table}

\underline{Loss Functions}\\
We use the mean square error measure as the loss function of the autoencoder and the negative log-likelihood as the loss function of the classifier. More precisely, we have
\begin{align}
L_{ae}&=\frac{1}{N} \sum_{j=1}^{N} \left(X_j - X'_j\right)^2\label{eqn:L_e2e}\\
L_\mathsf{C}&= - \sum_{j=1}^{M} Y_{\mathsf{NP},j} \log\left(W_{\mathsf{NP},j}\right) + \left(1-Y_{\mathsf{NP},j}\right) \log\left(1-W_{\mathsf{NP},j}\right),\label{eqn:L_C}
\end{align}
where $N$ and $M$ are the dimensions of the vectors $\mathbf{X}$ and $\mathbf{Y}_\mathsf{NP}$, respectively. Moreover, the negative log-likelihood loss function has been used to train the utility provider and adversary.

\underline{Training}\\
The training phase is explained in Algorithm \ref{alg:train}. A batch with the size of $n_b$ samples is taken from the dataset, and the parameters of all the DNNs $\mathsf{E}_\alpha$, $\mathsf{R}_\phi$, $\mathsf{C}_\theta$, and $\mathsf{D}_\beta$ are updated using the loss function defined in \eqref{eqn:L_e2e}. Then, keeping the networks $\mathsf{R}_\phi$ and $\mathsf{D}_\beta$ unchanged, we update the $\mathsf{E}_\alpha$ and $\mathsf{C}_\theta$ parameters using the loss function in \eqref{eqn:L_C} and repeat this for $n_e$ epochs. It should be noted that the training process is carried out without considering functions $f(\cdot)$ and $g(\cdot)$.
\begin{algorithm}[t!]
\caption{Training phase of the framework}
\textbf{Input:} Training dataset samples $\mathbf{X}$ \\
\hspace*{\algorithmicindent} parameter: learning rate $a$\\
\hspace*{\algorithmicindent} parameter: training steps $n_e$ and $n_b$ \\
\textbf{Output:} Obfuscator Model\\
\hspace*{\algorithmicindent} Initialization.
\begin{algorithmic}[1]
\For{$n_e$ epochs}
\State \parbox[t]{\dimexpr\textwidth-\leftmargin-\labelsep-\labelwidth}{Randomly select the mini-batch\\from the training data set.\strut}
\For{$i=0$ to $n_b$ iterations}
\State \parbox[t]{\dimexpr\textwidth-\leftmargin-\labelsep-\labelwidth}{Update the decoder parameters: \\ $ \beta_{i+1}=\beta_i - a\nabla_{\beta}L_{ae}(\beta_i,\mathbf{X}) $ \strut}
\State \parbox[t]{\dimexpr\textwidth-\leftmargin-\labelsep-\labelwidth}{Update $\mathsf{R}_\phi$ parameters: \\ $ \phi_{i+1}=\phi_i - a\nabla_{\phi}L_{ae}(\phi_i,\mathbf{X}) $ \strut}
\State \parbox[t]{\dimexpr\textwidth-\leftmargin-\labelsep-\labelwidth}{Update the classifier parameters: \\ $ \theta_{i+1}=\theta_i - a\nabla_{\theta}L_{ae}(\theta_i,\mathbf{X}) - a\nabla_{\theta}L_{\mathrm{C}}(\theta_i,\mathbf{X},\mathbf{Y}_{\mathsf{NP}})$ \strut}
\State \parbox[t]{\dimexpr\textwidth-\leftmargin-\labelsep-\labelwidth}{Update the encoder parameters: \\ $ \alpha_{i+1}=\alpha_i - a\nabla_{\alpha}L_{ae}(\alpha_i,\mathbf{X}) - a\nabla_{\alpha}L_{\mathsf{C}}(\alpha_i,\mathbf{X},\mathbf{Y}_{\mathsf{NP}})$ \strut}
\EndFor
\EndFor \\
\Return
\end{algorithmic}
\label{alg:train}
\end{algorithm}

As can be seen in Algorithm \ref{alg:train}, in each iteration, the weights are updated based on both the autoencoder and classifier loss functions, which makes not only the output of the autoencoder remain as similar as possible to the input data (maintain its utility), but also the output of the classifier (desirable feature) is well uncorrelated from the rest of the features.

\underline{Selection of Functions $f(\cdot)$ and $g(\cdot)$}\\
It is difficult to solve the optimization problem \eqref{eqn:DNN_func_Loss} to obtain optimal functions. Instead of optimally choosing these functions, here we introduce a natural and intelligent choice for them.

The function $g(\cdot)$ should be chosen to contributes to the dataset's utility for inferring non-private data. Therefore, a natural choice is to modify the classifier's output according to the labels in the original dataset. Assume that we consider smiling a feature required by the utility provider; since the LogSoftMax function is selected as the last layer of the classifier, the exponential value of its output represents the probability of smiling or not smiling, which can be between $0$ and $1$. Therefore, the function $g(\cdot)$ changes the output produced by the classifier to $\log(0)$ or $\log(1)$ depending on whether the image in the original dataset is smiling or not smiling. It should be noted that since $\log(0)$ cannot be used, we instead choose a sufficiently large negative value and call it $\lambda$. The reason for using LogSoftMax in the last layer of the classifier is that the output values of the classifier and the network $\mathsf{R}_\phi$ are in the same numerical range, and both are effective in the obfuscated output image. Also, the existence of the $\lambda$ parameter will effectively control the utility of the obfuscated dataset. Using SoftMax instead of LogSoftMax makes the obfuscator not converge or converge after a large number of epochs. In addition, using LogSoftMax has better numerical properties and makes training more stable \cite{pytorchdocssoftmax}.

The input to the function $f(\cdot)$ are the features we want to remain private and are well uncorrelated from the non-private features. Therefore, adding Gaussian noise is a suitable choice for the function $f(\cdot)$. The higher the amount of noise, the higher the level of privacy, and of course, the usefulness of the dataset is affected from the point of view of all features except non-confidential features. The effect of changing noise variance and $\lambda$ has been evaluated with simulations.

\underline{Utility Provider and Adversary Structures}\\
The structure of both of them is similar to the structure of the encoder plus classifier. The only difference will be in the output number of the last layer, which is proportional to the number of private features desired by the adversary or the number of features desired by the utility provider for inference.

\underline{Measurement of Utility and Privacy}\\
We assume that the adversary intends to infer the binary feature of gender (male/female) from the obfuscated dataset. The adversary is trained to infer this feature and then applied to the obfuscated dataset. Its accuracy in correctly diagnosing males or females is considered a measure of confidentiality.

The features of open mouth, smiling, and high cheekbone are considered as desirable features of the utility provider. We train the utility provider on the obfuscated dataset and then test the trained network on the original dataset. The accuracy of the utility provider in the desired feature recognition is considered a measure of utility.

As stated in the introduction, a significant drawback of the methods proposed for data privacy is that each data provider must separately train a network for this purpose. That puts a lot of burden on data providers. In our proposed method, the utility provider or a trusted authority can design and train an obfuscator according to the feature he wants to remain useful and share it among all data providers. Each data provider can then simply adjust the model's utility-privacy trade-off by adjusting the noise level and the value of $\lambda$ based on their sensitivity to their data.

\subsection{Categorical Dataset}

UCI Adult dataset includes census data of $48842$ people and $14$ categorical and continuous features of them \cite{dua2017uci}. Putting \cite{mandal2022uncertainty} as a benchmark, we convert the $14$ features into a vector of length $106$ by removing missing value data, normalizing the variables, and using one-hot encoding for categorical features. The binary features of income (more or less than $50$K per year) and gender are considered the demands of the utility provider and the adversary, respectively. In addition, the binary features of gender and income, which are adversary and utility provider preferences, are removed from the dataset. The resulting dataset with a feature vector of length $102$ is used for training networks.

\underline{DNNs Structures}\\
All DNNs comprise $3$ fully-connected layers, except $\mathsf{R}_\phi$, which comprises $2$ fully-connected layers. The details of the DNNs are given in Table \ref{tab:3}.
\begin{table}[t!]
\centering
\caption{DNNs architecture details for categorical datasets.}
\resizebox{\columnwidth}{!}{
\begin{tabular}{cccccc}
\hline 
\rule[-1ex]{0pt}{2.5ex} Component & Num & Layer & Output Size & Specs & Activation Function \\ 
\hline 
\rule[-1ex]{0pt}{2.5ex} Input Data &   & Samples & $102$ &   &   \\ 
\hline 
\rule[-1ex]{0pt}{2.5ex} Encoder & 1 & Linear & 128 &   & ReLU \\ 
\rule[-1ex]{0pt}{2.5ex}   & 2 & Linear & 128 &   & ReLU \\ 
\rule[-1ex]{0pt}{2.5ex}   & 4 & Linear & 64 &   & ReLU \\ 
\hline 
\rule[-1ex]{0pt}{2.5ex} Classifier & 1 & Linear & 32 &   & ReLU \\ 
\rule[-1ex]{0pt}{2.5ex}   & 2 & Linear & 8 &   & ReLU \\ 
\rule[-1ex]{0pt}{2.5ex}   & 4 & Linear & 2 &   & LogSoftMax \\ 
\hline 
\rule[-1ex]{0pt}{2.5ex} $\mathsf{R}_\phi$ & 1 & Linear & 64 &   & ReLU \\ 
\rule[-1ex]{0pt}{2.5ex}   & 2 & Linear & 62 &   & ReLU \\ 
\hline 
\rule[-1ex]{0pt}{2.5ex} Decoder & 1 & Linear & 128 &   & ReLU \\ 
\rule[-1ex]{0pt}{2.5ex}   & 2 & Linear & 128 &   & ReLU \\ 
\rule[-1ex]{0pt}{2.5ex}   & 4 & Linear & 102 &   & Sigmoid \\ 
\hline 
\rule[-1ex]{0pt}{2.5ex} Adversary & 1 & Linear & 256 & Dropout (p=0.2) & ReLU \\ 
\rule[-1ex]{0pt}{2.5ex} and & 2 & Linear & 256 & Dropout (p=0.3) & ReLU \\ 
\rule[-1ex]{0pt}{2.5ex} Utility Provider & 3 & Linear & 128 & Dropout (p=0.4) & ReLU \\ 
\rule[-1ex]{0pt}{2.5ex}   & 4 & Linear & 2 &   & LogSoftMax \\ 
\hline 
\end{tabular}
} 
\label{tab:3}
\end{table}

\underline{Utility Provider and Adversary Structures}\\
Utility provider and adversary are $4$-layer fully-connected networks that take an input of size $102$ and convert it into $2$ outputs (gender (male/female) for the adversary and income ($\gtrless 50$K) for the utility provider). Details are in Table \ref{tab:3}.

\section{Experiments}
\label{sec:simulation}

The proposed scheme is implemented in this section, and its performance is compared with other methods.
 
\subsection{Experiments Settings}
To compare and evaluate the performance, the settings (Input dataset, the size ratio of training, validation and test sets, etc.) used for image and categorical datasets are similar to \cite{singh2022decouple} and \cite{mandal2022uncertainty}, respectively. For the image dataset, the desirable features of the utility provider are open mouth, smiling, and high cheekbone, and the private feature is gender. For the categorical dataset, the utility provider is interested in inferring income (is it more or less than $50$K per year?), and the adversary is interested in understanding gender.

\underline{Dataset}\\
We choose CelebA as our image dataset. CelebA is a large-scale facial feature dataset with $202,599$ face images from $10,177$ identities and $40$ binary feature labels (such as gender, age, smile, etc.) in each image. $162,752$ samples were used as a training set, and the rest were used as a test and validation set. The obfuscator is trained on the training set and is validated in each epoch using the validation set. Finally, the performance of the network is evaluated on the test set. To learn the adversary, again, the CelebA dataset is used considering the gender feature as the data to be inferred while the utility provider training is done on the obfuscated dataset, and then the utility provider performance is tested on the non-obfuscated dataset.

UCI Adult has been selected as a categorical dataset containing the census data of $48842$ people. As explained in the previous section, with pre-processing corresponding to each record, we have a feature vector with length $106$. The desired features of the utility provider and the adversary are income and gender. $80\%$ of the dataset is used for training and the rest for testing.

\underline{Implementation Details}\\
Our experiment is implemented by PyTorch \cite{paszke2019pytorch} on Google Colab GPUs. The size of the images is $64\times64\times3$, and the mini-batch technique is used with a batch size of $n_b=64$. We use Adam optimizer \cite{kingma2014adam} to train all our networks and set the learning rate of all optimizers to $0.001$. Obfuscator, adversary, and utility provider training have been conducted with similar setups. The initial value of the weights for all networks is randomly generated with a Gaussian random variable with a variance of $0.02$ and an average of $0$ for the convolutional layers and an average of $1$ for the batch normalization layers. In addition, when the validation loss starts to increase, we stop the learning algorithm and use dropout with probability $0.5$ in some layers to protect the training from overfitting.

For the categorical dataset, we have $48842$ records, and corresponding to each record after one-hot encoding, there is a feature vector with length $106$. We use $39074$ records for training and the rest for testing. Also, we remove the features of income and gender, which are the demands of the utility provider and the adversary, respectively, from the dataset and use the resulting dataset for obfuscator training. Weights are initialized by Gaussian distribution with mean $0$ and variance $0.02$.

The implementation codes of the proposed scheme are available in \url{https://github.com/bozorgmehr77/adjustable-privacy}.

\subsection{Experiments Results}
\begin{figure}[t!]
\centering
\captionsetup{justification=centering,margin=2cm}
\includegraphics[scale=0.85]{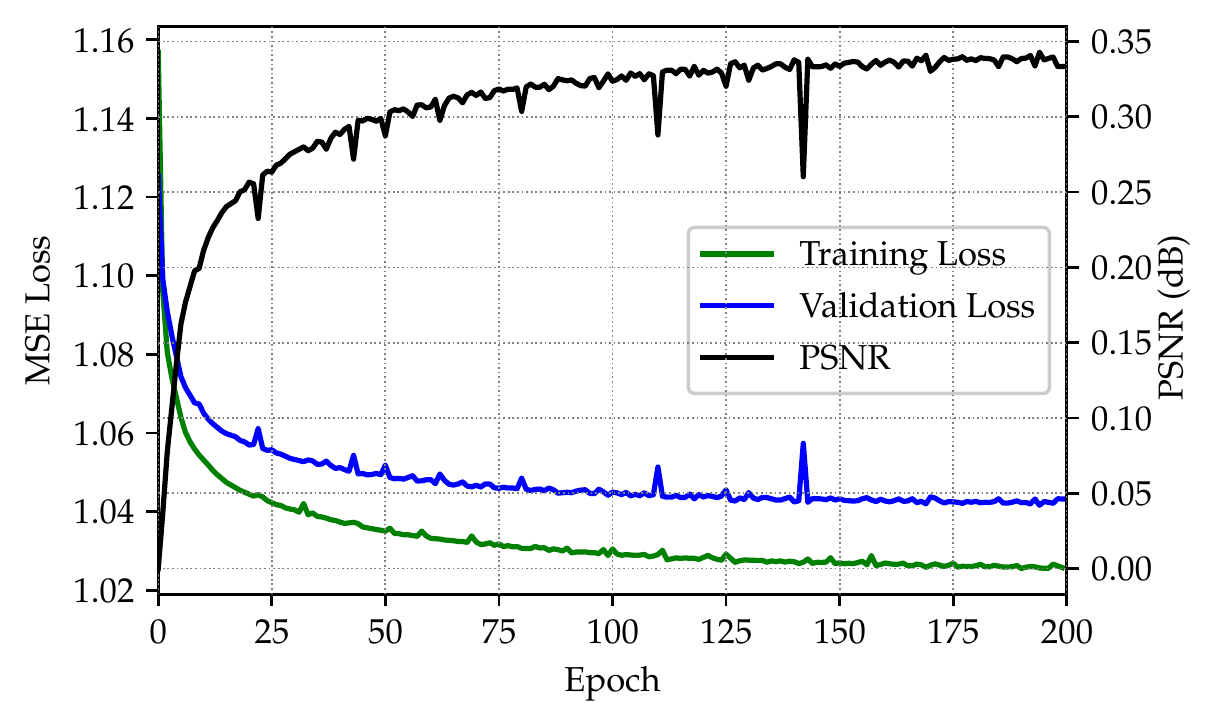}
\caption{Training and validation losses related to autoencoder.}
\label{fig:e2e_psnr_and_loss_plot}
\end{figure}

Let's assume smiling is the desire of the utility provider, and gender is the adversary's desire. By training the obfuscator on the CelebA dataset for several epochs, the validation and training errors of the autoencoder are plotted in Fig. \ref{fig:e2e_psnr_and_loss_plot}. As can be seen from Fig. \ref{fig:e2e_psnr_and_loss_plot}, with the increase in the number of epochs, the training and validation errors related to the autoencoder are both reduced and as a result, the obfuscated image $\mathbf{X}'$ is getting closer to the original image $\mathbf{X}$. Therefore, according to Fig. \ref{fig:e2e_psnr_and_loss_plot}, the PSNR between $\mathbf{X}$ and $\mathbf{X}'$ increases. Reducing the autoencoder error makes the obfuscated image remain as useful as possible despite the compression, while the classifier helps to preserve the desired feature (smiling here) well in the final dataset. Therefore, controlling the errors of the autoencoder preserves the desired feature and makes it well uncorrelated from other features. Later, in subsection \ref{subsec:epoch_eff}, we will discuss the validation of the obfuscator training process.

After the training phase, we activate the functions $f(\cdot)$ and $g(\cdot)$. The function $g(\cdot)$ converts the output of the classifier to $\log(0)\approx\lambda$ and $\log(1)=0$ depending on the smiling feature in the original dataset, and the function $f(\cdot)$ is an additive white Gaussian noise (AWGN) with $\mu=0$ and variance proportional to $\nu$. Here, $\nu$ is the average of the $\mathsf{R}$ output nodes for each record.

Suppose the model corresponding to the epoch number $183$ is selected, the noise variance is set to $60\nu$, and $\lambda=-3000$. Then we convert the whole dataset $\mathcal{D}$ to dataset $\mathcal{D}'$ by passing through obfuscator. Now the adversary infers the gender from the obfuscated dataset, and its accuracy in recovering the gender, which is our privacy criterion, in weak and strong cases are $\%53.55$ and $\%63.24$, respectively, while the accuracy of gender inference from the main dataset is equal to $\%97.30$. Therefore, the proposed algorithm has strengthened privacy in weak and strong adversarial cases by $\%43.75$ and $\%34.06$, respectively.

In terms of utility, assuming that the utility provider is trained with the obfuscated dataset, its inference accuracy for the smiling feature from the original dataset is equal to $\%85.43$. But if it is trained on the main dataset, this value will equal $\%91.92$. This slight decrease of $\%6.49$ in usefulness shows the strength of the proposed method.

\subsubsection{Effect of Noise Variance on Utility-Privacy Trade-off}
\begin{figure}[t!]
\centering
\captionsetup{justification=centering}
\includegraphics[width=0.9\columnwidth]{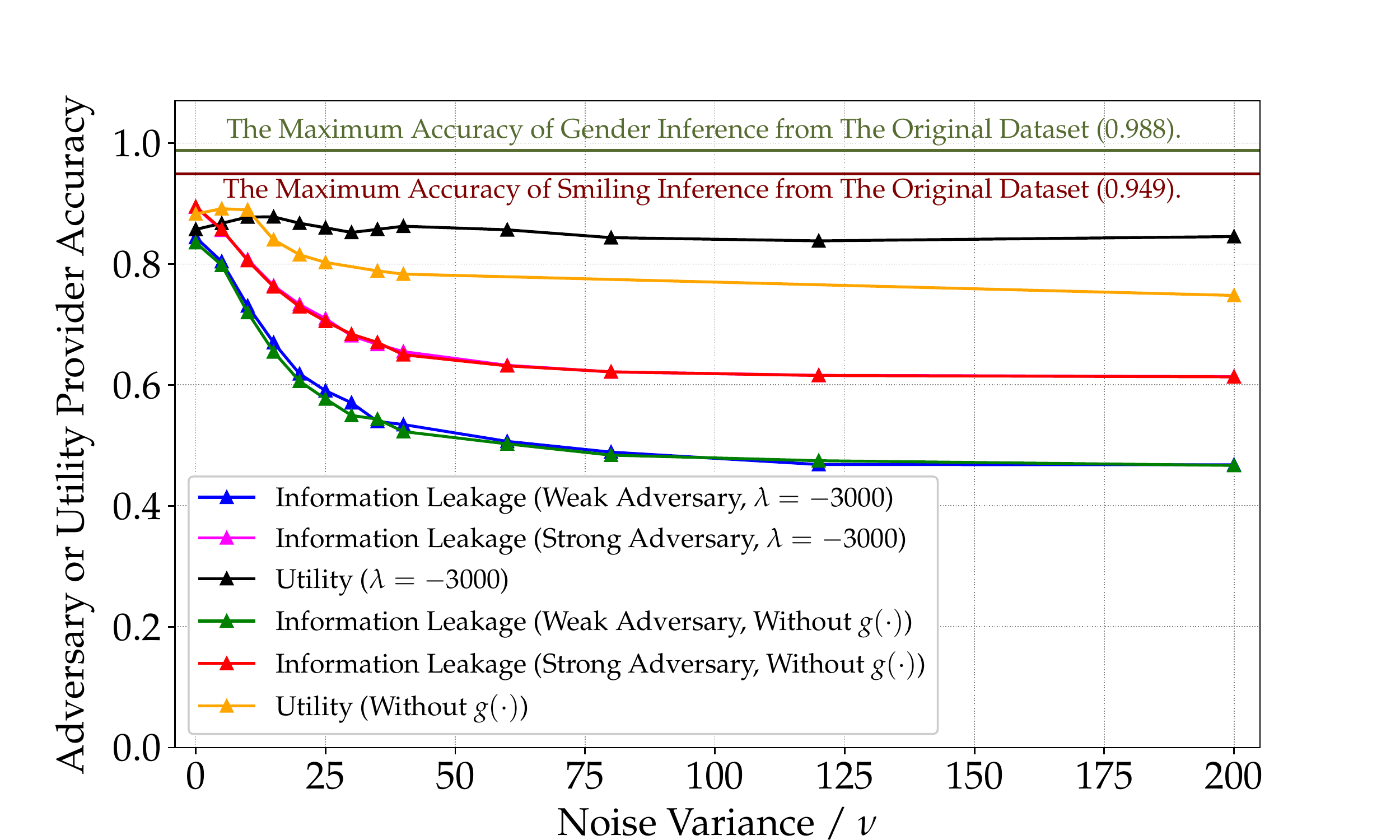}
\caption{Utility and privacy in terms of the increase of noise variance for the CelebA-G-S and considering the cases where the function $g(\cdot)$ is active or inactive.}
\label{fig:UorP_vs_noise_plot}
\end{figure}

The noise variance is a parameter the data provider can use to adjust privacy. Considering the CelebA-G-S case, we train the obfuscator for $200$ epochs. We set $\lambda$ equal to $-3000$ and change the noise variance in the interval $[0,200\nu]$. Fig. \ref{fig:UorP_vs_noise_plot} shows the changes in privacy and utility with noise variance. This figure gives the results for weak and strong adversaries and the case where the function $g(\cdot)$ is not applied. The following points are evident from Fig. \ref{fig:UorP_vs_noise_plot}.

\begin{itemize}
\item By increasing the noise, the utility increases slightly first because adding a small amount of noise makes the dataset more diverse. As a result, the utility provider is better trained on it. With the further increase of the noise and in the presence of the function $g(\cdot)$, the utility decreases slightly, while without applying $g(\cdot)$, the decrease is more significant. Therefore, the presence of the function $g(\cdot)$ preserves the dataset's utility regarding the smiling feature.

\item Comparing the utility and privacy curves for the case where the $g(\cdot)$ function is disabled, and we don't have noise (the starting point of the curves) with the maximum utility and privacy values shows that the privacy and utility have dropped by $0.066$ and $0.093$, respectively. The lower amount of utility loss is because the presence of the classifier makes the feature of smiling (non-confidential) better preserved.

\item For both strong and weak adversaries, privacy is strengthened by increasing the noise variance until it reaches the saturation limit. The saturation value for a weak adversary is about $50\%$, which is the same as a random guess, and for a strong adversary, it is $60\%$.

\item The comparison of privacy for two cases of active and inactive $g(\cdot)$ function shows that $g(\cdot)$ does not have much effect on privacy.
\end{itemize}

\begin{figure}
\centering
\captionsetup{justification=centering}
\includegraphics[width=0.9\columnwidth]{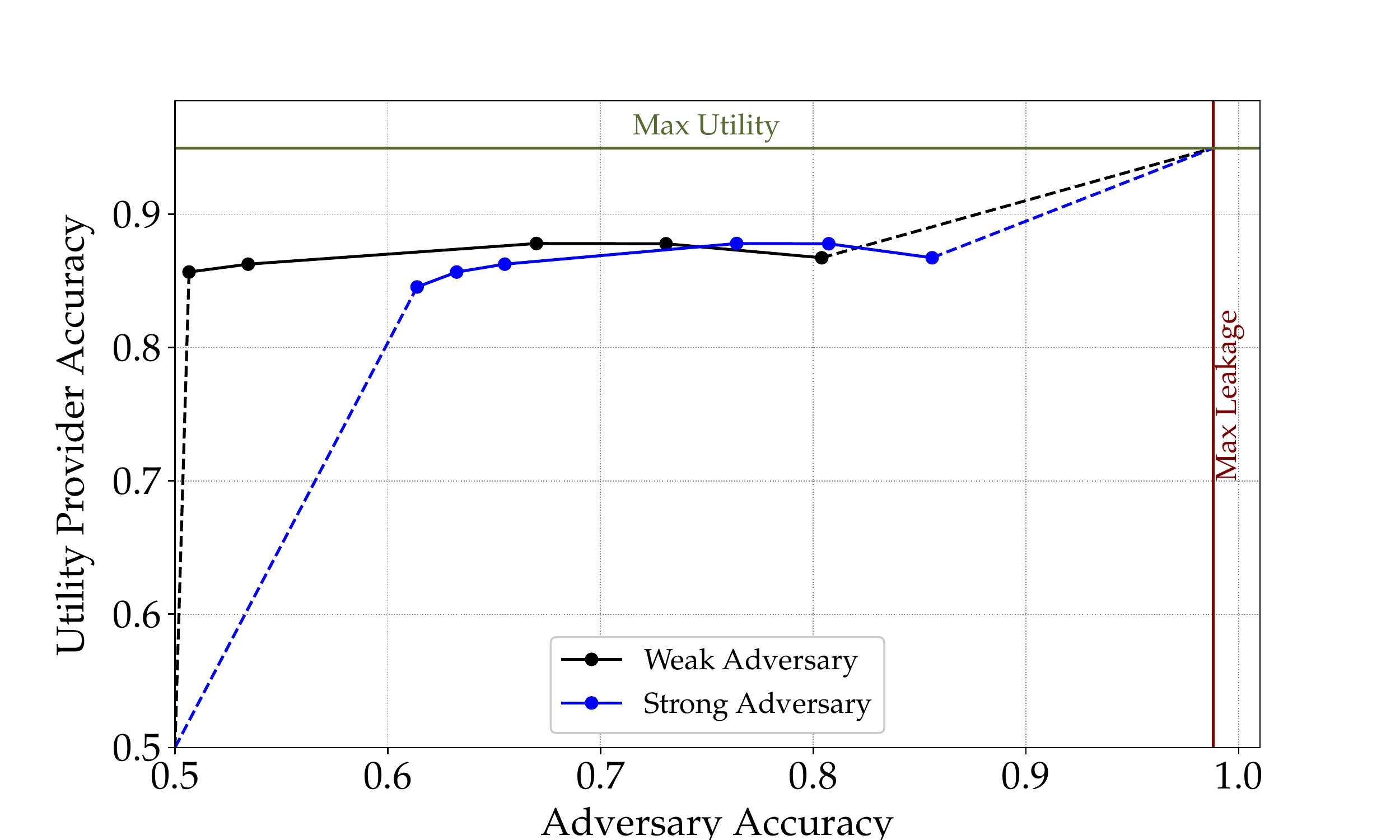}
\caption{Utility-privacy trade-off for case CelebA-G-S with strong and weak adversaries.}
\label{fig:strong_adv_plot}
\end{figure}

Moreover, the utility-privacy trade-off curves for strong and weak adversaries are depicted in Fig. \ref{fig:strong_adv_plot}. The horizontal axis represents information leakage as a measure of privacy, and the vertical axis represents utility. The ideal point is $(0.5,1)$; the closer the curve is to this point; the better the algorithm is from the point of view of the utility-privacy trade-off. It can be seen that the curve has shifted to the right for the strong adversary.

\subsubsection{Effect of \texorpdfstring{$\lambda$}{lambda} on Utility-Privacy Trade-off}
\begin{figure}
\centering
\captionsetup{justification=centering}
\includegraphics[width=0.9\columnwidth]{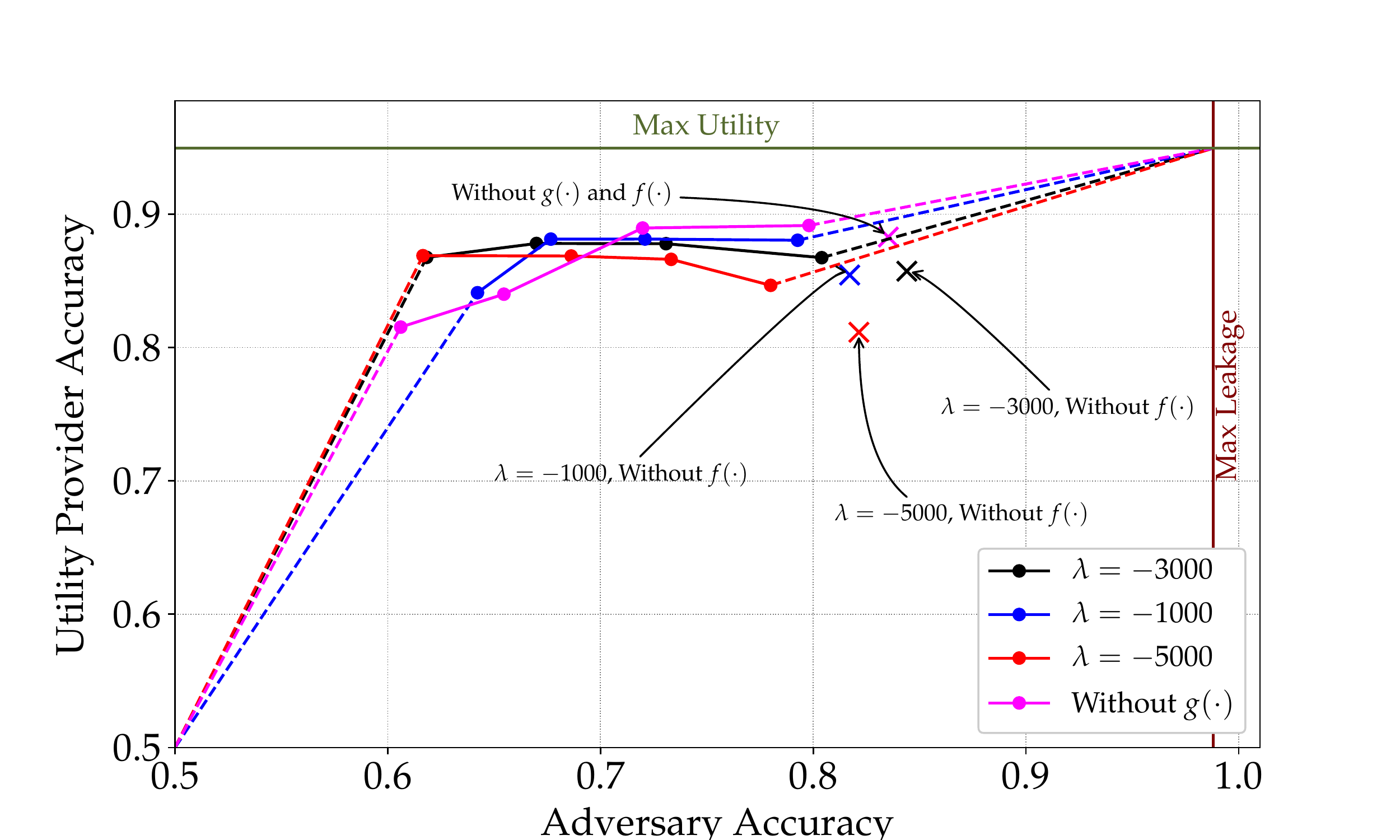}
\caption{Comparison of the utility-privacy trade-off for case CelebA-G-S and different values of $\lambda$.}
\label{fig:g_plot}
\end{figure}

$\lambda$ is a parameter that affects the utility of the dataset. Considering the case CelebA-G-S, the utility-privacy trade-off for $\lambda\in\{-5000,-3000,-1000\}$, the case where the function $g(\cdot)$ is not applied, and the noise variances of $\{0,5,10,15,20\}\nu$ are plotted in Fig. \ref{fig:g_plot}. The adversarial type is weak, and points related to not adding noise are marked with a cross. Further, the maximum utility and information leakage shown in the figure corresponds to the accuracy of inferring smiling and gender features by the utility provider and the adversary, respectively. The following points can be deduced from Fig. \ref{fig:g_plot}.
\begin{itemize}
\item By increasing the value of $|\lambda|$, the effect of the output value of the classifier increases. As a result, the power of inferring the desired feature of the utility provider increases in high noise. On the other hand, the diversity of the dataset is reduced, and the effect of other features fades, which leads to a small increase in privacy. By decreasing the value of $|\lambda|$, the inference power of the utility provider decreases and, consequently, the utility. Therefore, to maintain the dataset's diversity and the accuracy of the utility provider's inference, the value of $\lambda$ should be well adjusted.
\item Increasing the value of $|\lambda|$ reduces the diversity of the dataset, and in low noise variances, it leads to a slight decrease in utility.
\item With the addition of noise, the utility generally decreases. However, for $\lambda=-3000$, the utility is almost constant for a relatively wide range of noise variances, which makes it a suitable candidate.

\end{itemize}

\begin{figure}[t!]
\centering
\resizebox{.88\columnwidth}{!}{
\begin{tabularx}{\linewidth}{lX}
\rule[-1ex]{0pt}{2.5ex} Original Dataset & \includegraphics[scale=0.333]{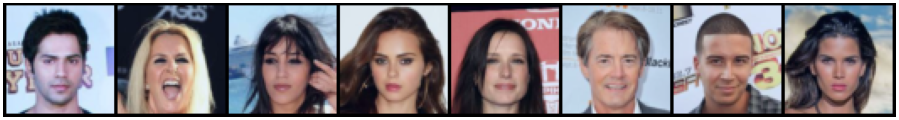} \\ 
\rule[-1ex]{0pt}{2.5ex} $\mathrm{Var}=10\nu$ and without $g(\cdot)$ & \includegraphics[scale=0.25]{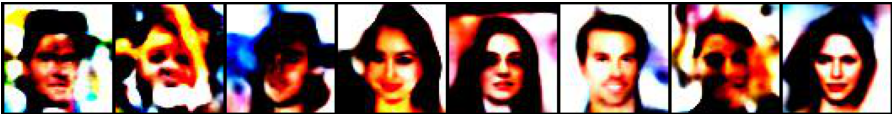} \\ 
\rule[-1ex]{0pt}{2.5ex} $\mathrm{Var}=10\nu$ and $\lambda=-3000$ & \includegraphics[scale=0.25]{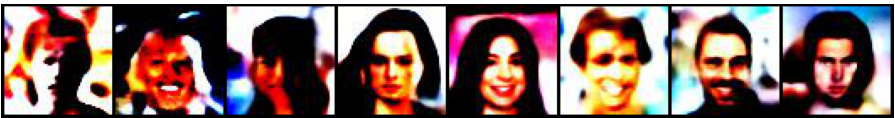} \\ 
\rule[-1ex]{0pt}{2.5ex} $\mathrm{Var}=20\nu$ and $\lambda=-3000$ & \includegraphics[scale=0.25]{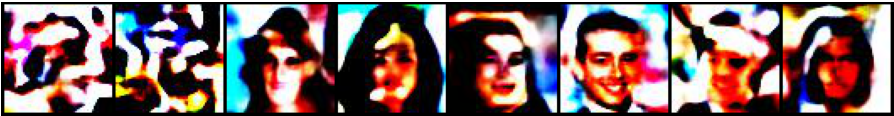} \\ 
\rule[-1ex]{0pt}{2.5ex} $\mathrm{Var}=10\nu$ and $\lambda=-5000$ & \includegraphics[scale=0.25]{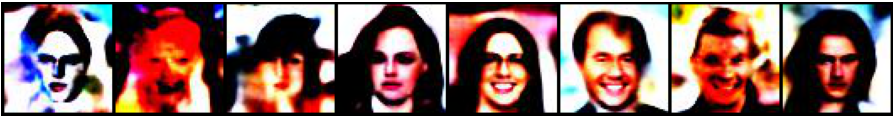} \\ 
\rule[-1ex]{0pt}{2.5ex} $\mathrm{Var}=20\nu$ and $\lambda=-5000$ & \includegraphics[scale=0.25]{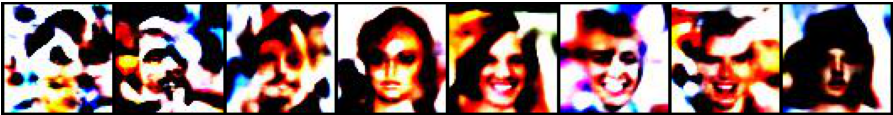} \\ 
\end{tabularx}
}
\caption{Obfuscated images using different values of noise variance and $\lambda$.}
\label{tab:pictures}
\end{figure}

Considering the case CelebA-G-S, the effect of different amounts of noise and $\lambda$ is given visually in Fig. \ref{tab:pictures}.

\subsubsection{Effect of Epoch Number on Utility-Privacy Trade-off}
\label{subsec:epoch_eff}

\begin{figure}[t!]
\centering
\captionsetup{justification=centering,margin=2cm}
\includegraphics[scale=0.9]{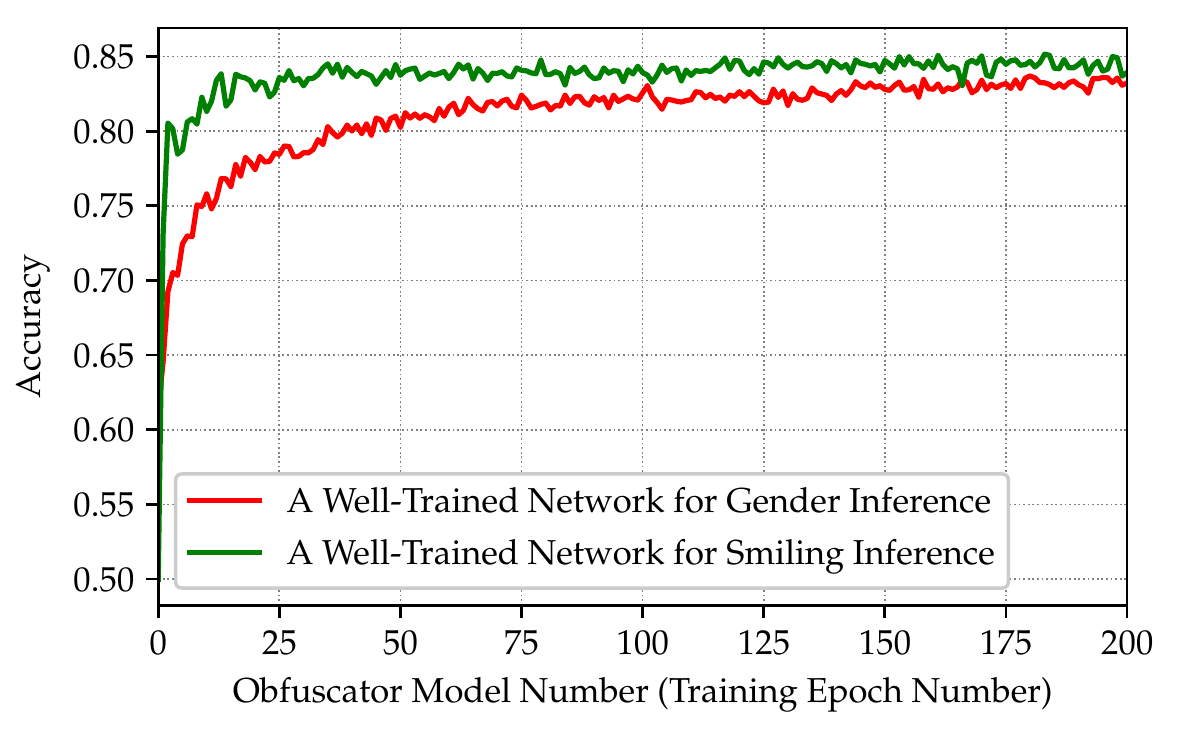}
\caption{Accuracy of well-trained networks for extracting gender and smiling features by the number of epochs.}
\label{fig:p2rNumber_selection_gs_plot}
\end{figure}

As seen in Fig. \ref{fig:e2e_psnr_and_loss_plot}, the autoencoder error continuously decreases with increasing epochs. However, to validate the training process, the overall performance of the obfuscator should be evaluated, and the epoch number should be determined based on this overall performance to have a satisfactory model. For this purpose, ignoring the functions $f(\cdot)$ and $g(\cdot)$, the inference accuracy of smiling and gender features by the well-trained networks is plotted in Fig. \ref{fig:p2rNumber_selection_gs_plot}. As you can see, with the increase in the number of epochs, both curves climb to a high value, which shows that both the utility provider has a good performance and the dataset remains diverse in terms of features other than non-private ones. Therefore, by increasing the number of epochs, the overall performance of the obfuscator will be better.

Consider the case CelebA-G-S and choose the models corresponding to the epoch numbers $13$ and $183$. According to Fig. \ref{fig:p2rNumber_selection_gs_plot}, in epoch $13$, the accuracy of inferring smiling is high, while the accuracy of gender inference is low. Herefore, the feature of smiling is well preserved while other features are somewhat degraded. In epoch $183$, where the autoencoder is well trained (based on Fig. \ref{fig:e2e_psnr_and_loss_plot}), the accuracy of both smiling and gender inference is high, indicating that both the classifier is well trained, and diversity of the obfuscated dataset is preserved.

\begin{figure}[t!]
\centering
\captionsetup{justification=centering}
\includegraphics[width=0.9\columnwidth]{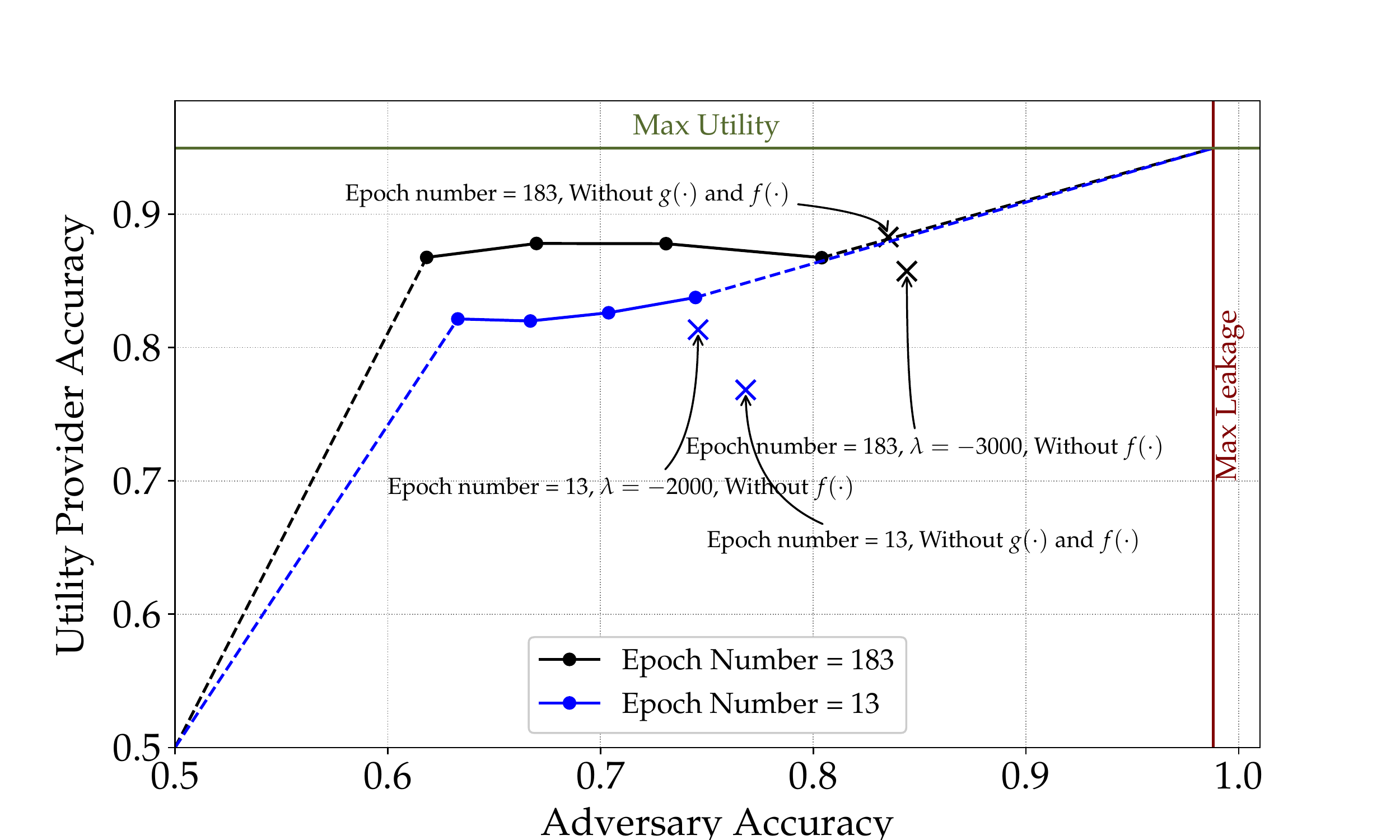}
\caption{Comparison of utility-privacy trade-off for obfuscator corresponding to models No. $13$ and $183$.}
\label{fig:p2r_plot}
\end{figure}

Suppose $\lambda$ is $-3000$ and $-2000$ for epochs $13$ and $183$, respectively, and we have a weak adversary; the utility-privacy trade-off is plotted in Fig. \ref{fig:p2r_plot} for epochs $13$ and $183$.  As you can see in the figure, more training has led to an increase in utility and a slight decrease in privacy. The reason for the reduction of privacy is that by reducing the autoencoder error, the output image preserves as many of the features of the input image as possible. As a result, the private feature is also present in the obfuscated dataset with higher quality. Regarding the utility, with less training of the autoencoder, the resulting dataset has less diversity, which will lead to less utility. Therefore, the model corresponding to a higher epoch performs better regarding the utility-privacy trade-off, and the obfuscator is not overfitted until at least epoch $200$.

\begin{figure}[t!]
\centering
\captionsetup{justification=centering,margin=2cm}
\includegraphics[scale=0.9]{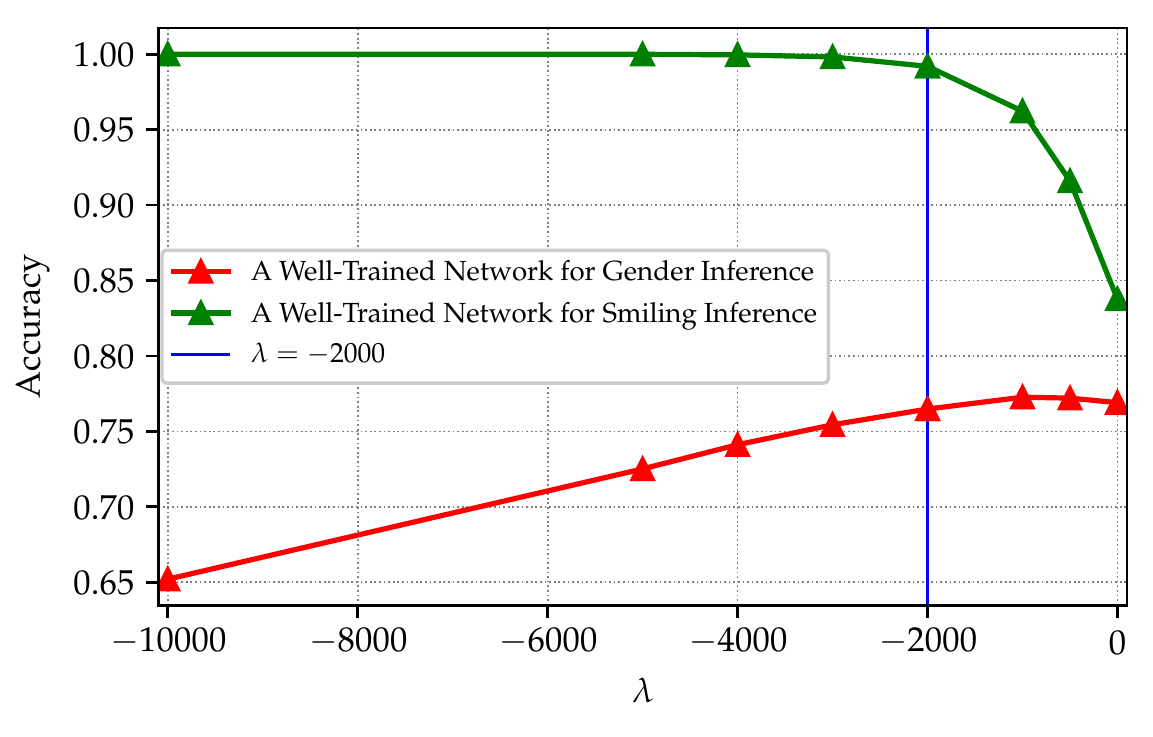}
\caption{Choosing a suitable $\lambda$ for obfuscator corresponding to the model number $13$ and case CelebA-G-S.}
\label{fig:g_selection_gs13_plot}
\end{figure}

\begin{figure}[t!]
\centering
\captionsetup{justification=centering,margin=2cm}
\includegraphics[scale=0.9]{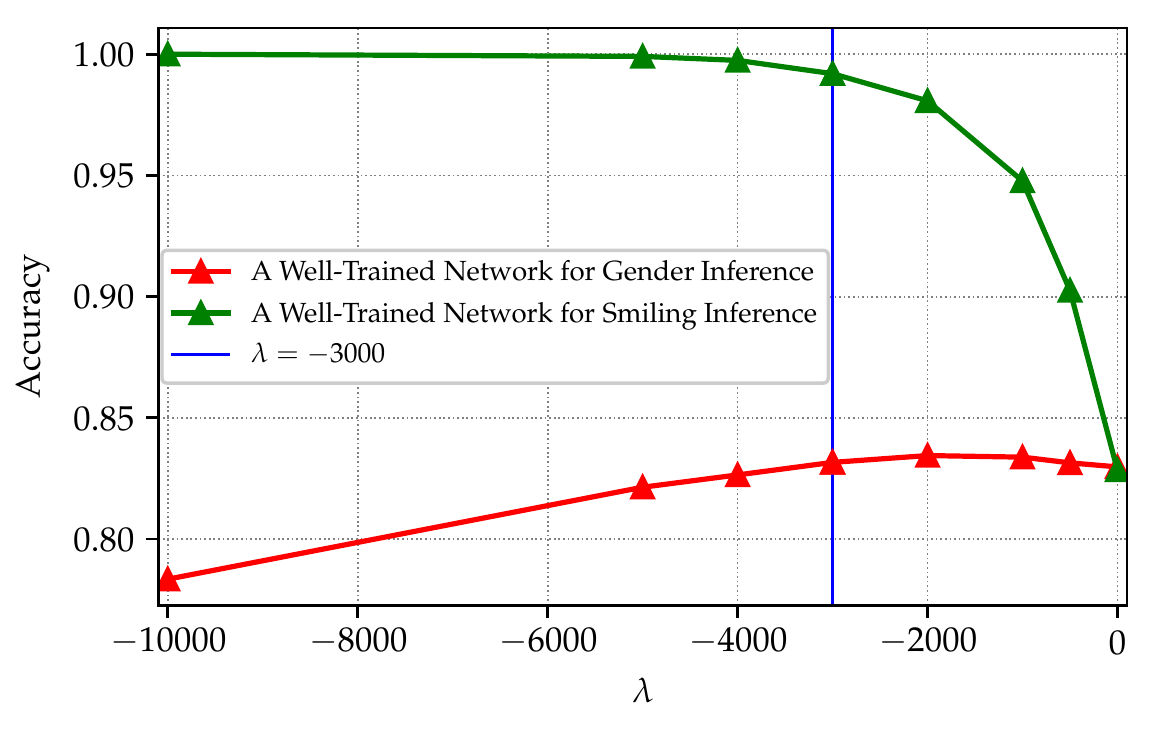}
\caption{Choosing a suitable $\lambda$ for obfuscator corresponding to the model number $183$ and case CelebA-G-S.}
\label{fig:g_selection_gs183_plot}
\end{figure}

It should be noted that the values of $\lambda=-2000,-3000$ have been chosen so that the models related to epochs $13$ and $183$ perform well, as shown in Figs. \ref{fig:g_selection_gs13_plot} and \ref{fig:g_selection_gs183_plot}. To depict these two figures, assuming that the function $f(\cdot)$ is inactive, the accuracy of two well-trained networks for inferring smiling and gender features has been evaluated in terms of the $\lambda$ parameter on the obfuscated test set. The optimal $\lambda$ point, as shown in the figure, is where both networks have reached high accuracy, and the accuracy of the gender-inferring network has not dropped too much to have diversity in the constructed data set. Here, gender represents all the features except non-private ones.

\subsubsection{Decorrelation}
\begin{figure}
\centering
\captionsetup{justification=centering}
\includegraphics[width=0.9\columnwidth]{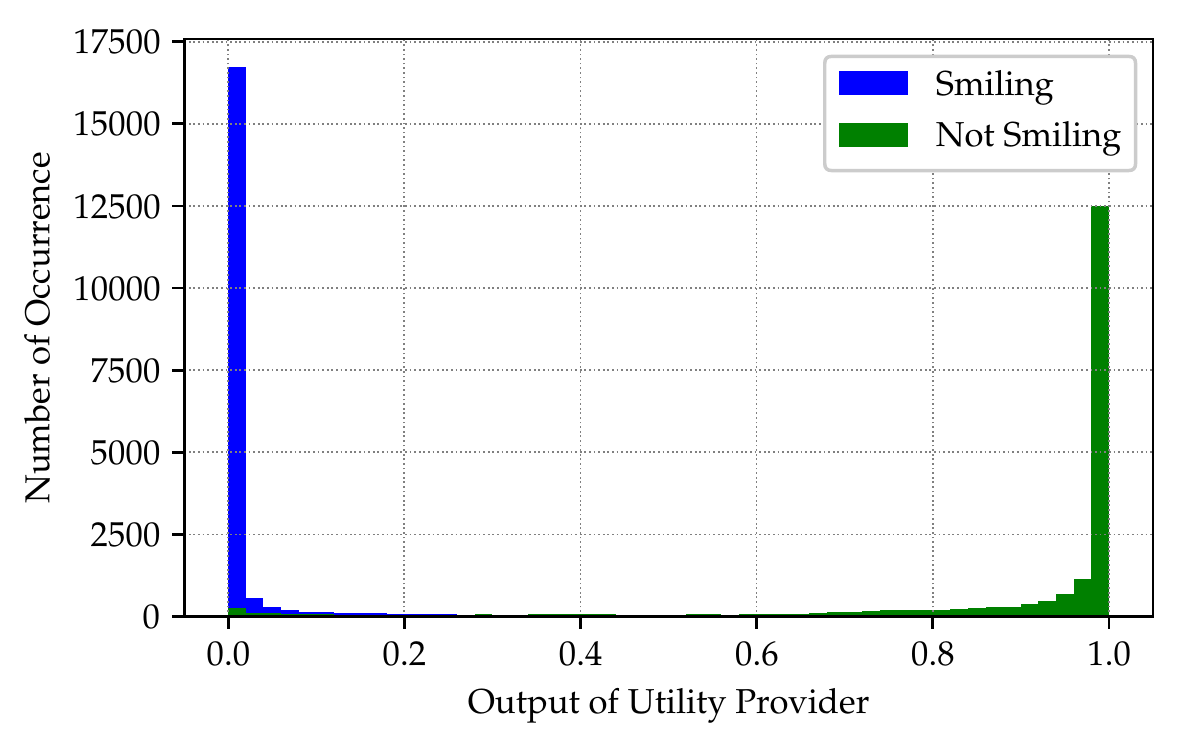}
\caption{Histogram of the utility provider prediction when the classifier's output is randomly labeled smiling or not smiling.}
\label{fig:hist_utl}
\end{figure}

In this section, we show that the proposed obfuscator makes the utility provider's desired feature well uncorrelated from the rest of the features. For this purpose, consider the case CelebA-G-S and set $\lambda$ to $-3000$. We also assume that the function $f(\cdot)$ is not applied. For each image, we randomly change the output of the classifier to $0$ or $-3000$ (smiling or not) and give it to the decoder along with the $\mathsf{R}_\phi$ output vector without applying noise to generate a new image. The output image is labeled smiling or not smiling by the utility provider. The histogram of inferring the smiling feature by the utility provider is given in Fig. \ref{fig:hist_utl}. As you can see, the images are well separated regarding smiling or not smiling. This shows that the classifier's output has almost complete control over the smiling feature, and the $\mathsf{R}_\phi$ output has almost no effect on the inference of this feature.

\subsubsection{Comparison with Similar Works}
\begin{figure}
\centering
\captionsetup{justification=centering}
\includegraphics[width=0.9\columnwidth]{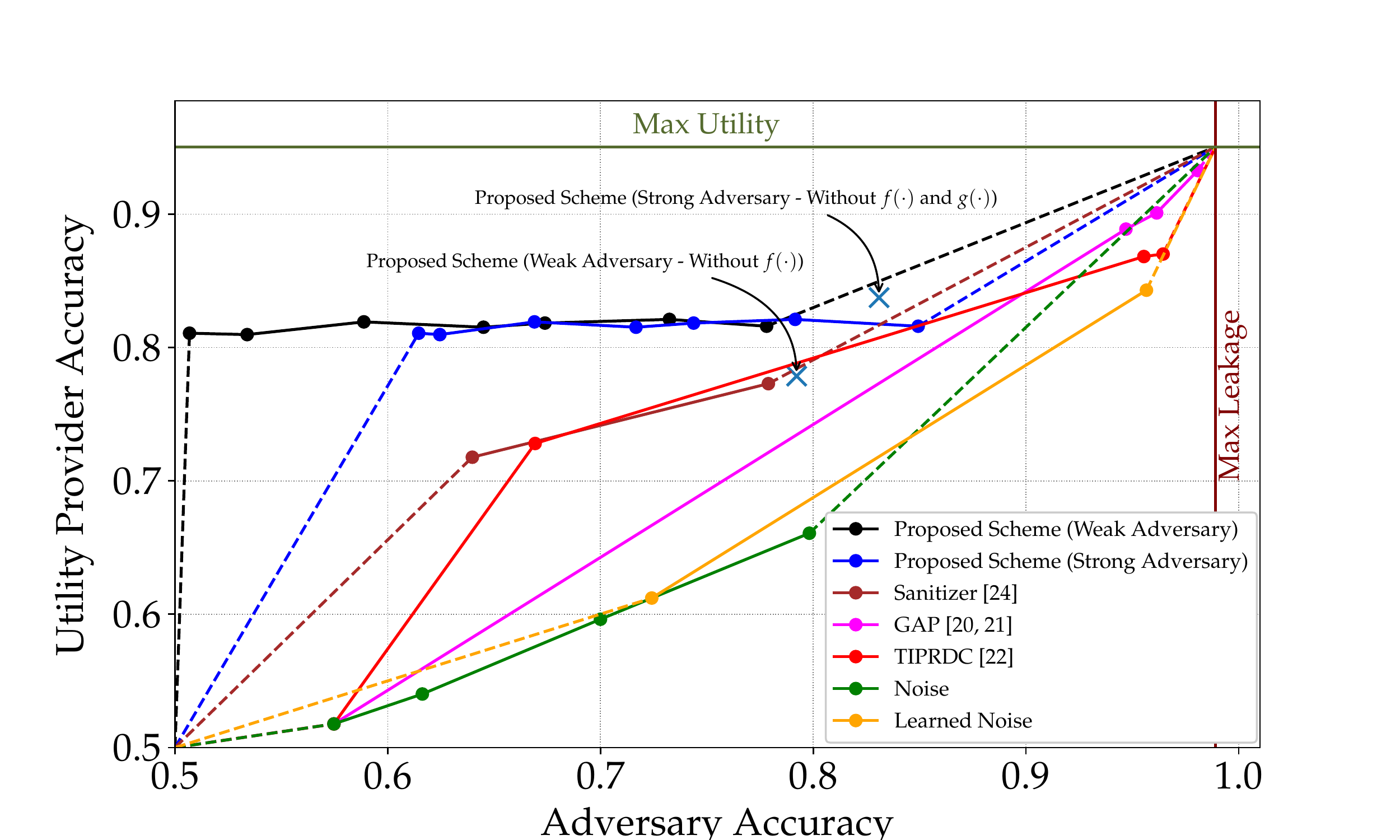}
\caption{Comparison of utility-privacy trade-off for different methods for case CelebA-G-M.}
\label{fig:UC1_UP_Tradeoff_plot}
\end{figure}

\begin{figure}
\centering
\captionsetup{justification=centering}
\includegraphics[width=0.9\columnwidth]{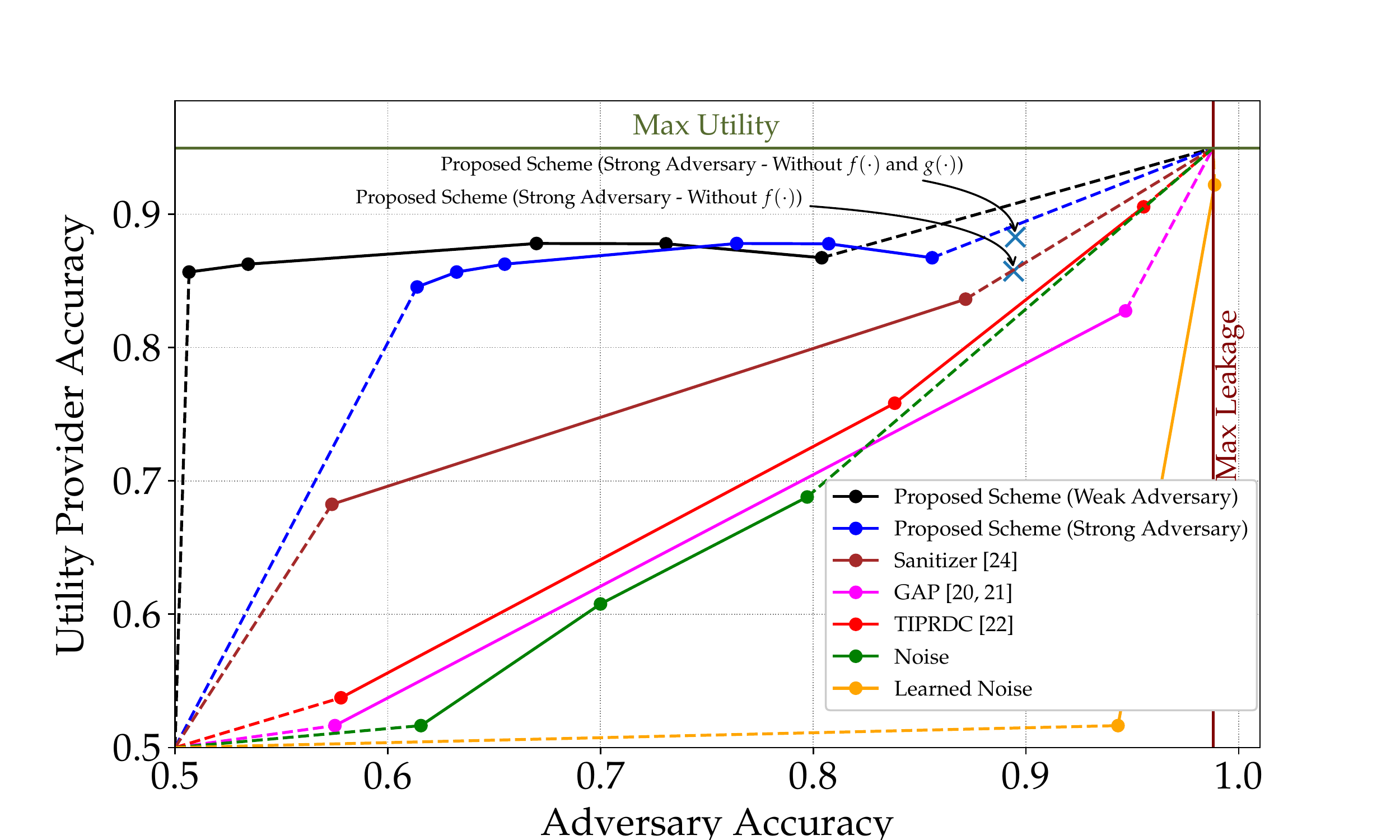}
\caption{Comparison of utility-privacy trade-off for different methods for case CelebA-G-S.}
\label{fig:UC2_UP_Tradeoff_plot}
\end{figure}

\begin{figure}
\centering
\captionsetup{justification=centering}
\includegraphics[width=0.9\columnwidth]{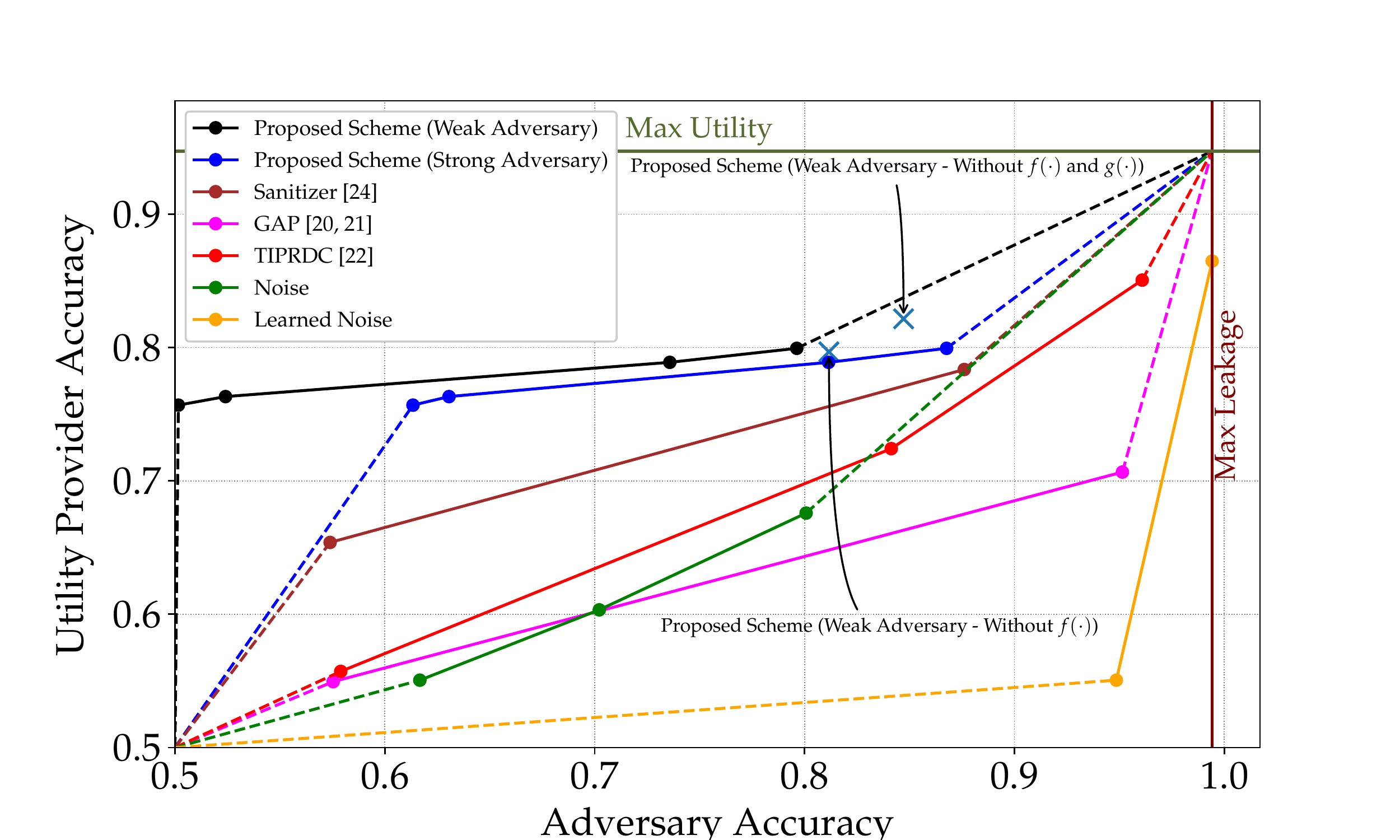}
\caption{Comparison of utility-privacy trade-off for different methods for case CelebA-G-C.}
\label{fig:UC3_UP_Tradeoff_plot}
\end{figure}

In this section, we have compared the proposed scheme with the methods of different references regarding the utility-privacy trade-off. For this purpose, the results obtained in \cite{singh2022decouple} have been used. This comparison has been made for all three cases of CelebA-G-M, CelebA-G-S, and CelebA-G-C and for both weak and strong adversaries. As is evident from Figs. \ref{fig:UC1_UP_Tradeoff_plot}, \ref{fig:UC2_UP_Tradeoff_plot} and \ref{fig:UC3_UP_Tradeoff_plot}, the proposed model outperforms other methods. The references of the algorithms used for comparison are mentioned in the figure. ``Noise" refers to adding Gaussian noise with zero mean and variance $40$ as done in \cite{li2020tiprdc}. In the learned noise method, first, a noise gets into a DNN, and the output is added to the dataset \cite{huang2017context,kairouz2022generating}. To make a fair comparison, the adversary and the utility provider are designed to be almost identical to the previous works regarding maximum privacy and utility. Another positive point about the proposed scheme is that utility provider converges in low epochs, generally less than $10$. Strong adversaries are considered in the case of curves related to previous work. In addition, the utility provider is trained on the obfuscated dataset, and its accuracy is checked on the obfuscated dataset, while the logical assumption is that the accuracy should be calculated on the original dataset (similar to what we did in this paper). As you can see in the figure, the performance curve will be better assuming the same utility provider as the previous works.

\subsubsection{Categorical Dataset}

\begin{figure}
\centering
\includegraphics[width=0.9\columnwidth]{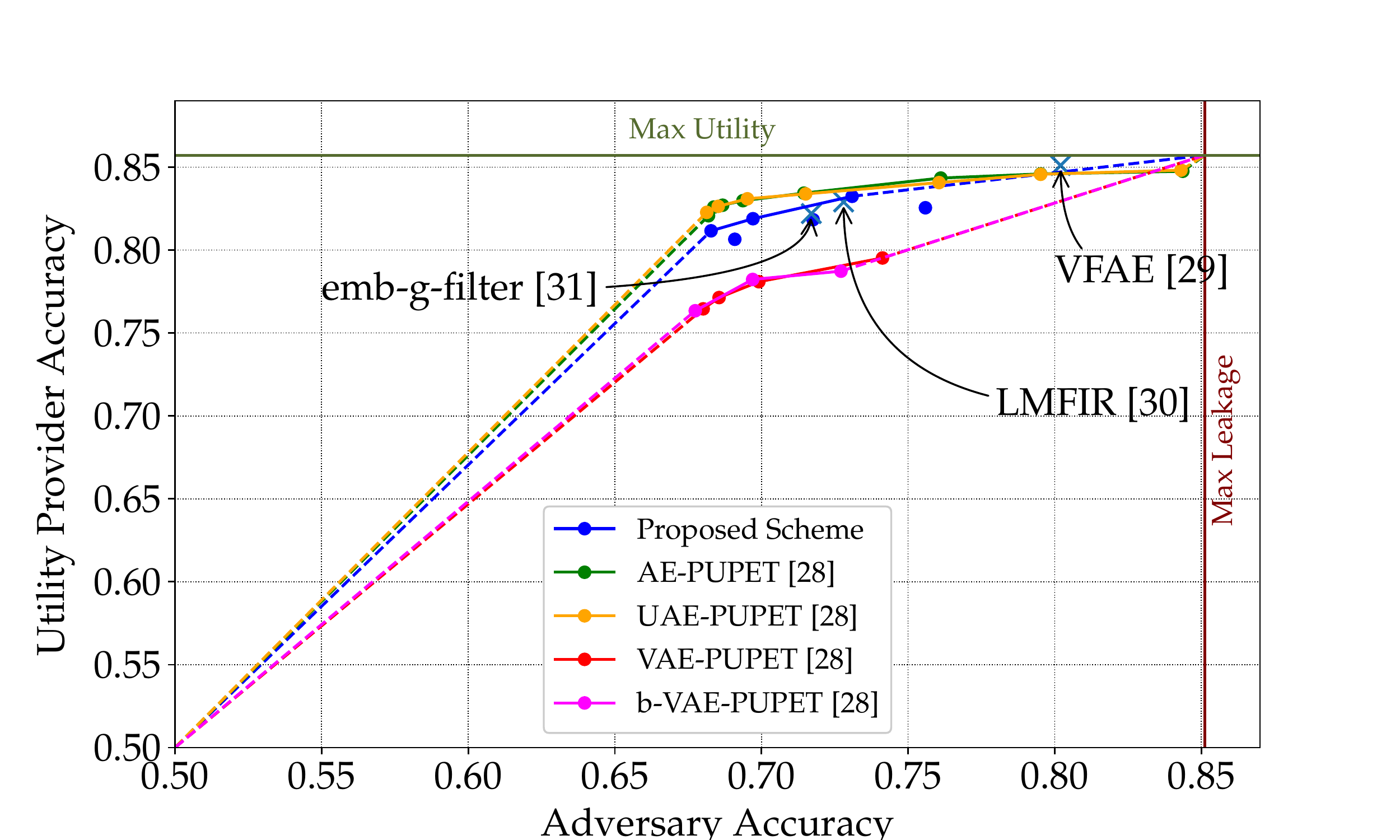}
\caption{Comparison of utility-privacy trade-off for different methods on UCI Adult dataset. The desirable features of the adversary and the utility provider are gender and income, respectively.}
\label{fig:UCI_1_UP_Tradeoff_plot}
\end{figure}

For the categorical dataset, the performance of the proposed algorithm is compared with the previous methods in terms of utility-privacy trade-off in Fig. \ref{fig:UCI_1_UP_Tradeoff_plot}. All curve values of previous works are taken from \cite{mandal2022uncertainty}. AE-PUPET, UAE-PUPET, VAE-PUPET, and b-VAE-PUPET methods are related to \cite{mandal2022uncertainty}, and VFAE, LMFIR, and emb-g-filter methods are introduced in \cite{louizos2015variational}, \cite{song2019learning}, and \cite{chen2019distributed}, respectively.

\subsubsection{Area Under Curve}

Calculating the area under the curve (AUC) can be a good measure to compare different algorithms regarding the utility-privacy trade-off. For this purpose, we consider the convex hull of the curve and then calculate AUC \cite{mandal2022uncertainty}. AUC values for different schemes are compared in Table \ref{tab:4}. The AUC of the proposed scheme for the image dataset is more than other methods, and it competes with other methods in the case of the categorical dataset. For comparison, to calculate AUC similar to \cite{singh2022decouple}, the origin point is considered $(0,0)$, while it is logical to take the point $(0.5,0.5)$ as the origin.
\begin{table}[t]
\centering
\resizebox{\columnwidth}{!}{
\begin{tabular}{l|c|c|c}
\hline 
\rowcolor{lightgray} \rule[-1ex]{0pt}{2.5ex} Image Dataset & (CelebA-G-M) Convex Hull AUC & (CelebA-G-S) Convex Hull AUC & (CelebA-G-C) Convex Hull AUC \\ 
\hline 
\rule[-1ex]{0pt}{2.5ex} Sanitizer \cite{singh2022decouple} & $0.5210$ & $0.5337$ & $0.5241$ \\ 
\hline 
\rule[-1ex]{0pt}{2.5ex} TIPRDC \cite{li2020tiprdc} & $0.5121$ & $0.4690$ & $0.4736$ \\ 
\hline 
\rule[-1ex]{0pt}{2.5ex} GAP \cite{huang2017context,kairouz2022generating} & $0.4701$ & $0.4690$ & $0.4715$ \\ 
\hline 
\rule[-1ex]{0pt}{2.5ex} Gaussian Noise & $0.4701$ & $0.4690$ & $0.4709$ \\ 
\hline 
\rule[-1ex]{0pt}{2.5ex} Learned Noise & $0.4701$ & $0.4693$ & $0.4709$ \\ 
\hline 
\rule[-1ex]{0pt}{2.5ex} Proposed Scheme & $0.5790$ & $0.5963$ & $0.5566$ \\ 
\hline 
\rowcolor{lightgray} \rule[-1ex]{0pt}{2.5ex} Categorical Dataset & Convex Hull AUC &   &   \\
\hline 
\rule[-1ex]{0pt}{2.5ex} AE-PUPET \cite{mandal2022uncertainty} & $0.4236$ \\ 
\hline 
\rule[-1ex]{0pt}{2.5ex} UAE-PUPET \cite{mandal2022uncertainty} & $0.4234$ \\ 
\hline 
\rule[-1ex]{0pt}{2.5ex} VAE-PUPET \cite{mandal2022uncertainty} & $0.3995$ \\ 
\hline 
\rule[-1ex]{0pt}{2.5ex} b-VAE-PUPET \cite{mandal2022uncertainty} & $0.4001$ \\ 
\hline 
\rule[-1ex]{0pt}{2.5ex} Proposed Scheme & $0.4183$ \\  
\hline 
\end{tabular} 
}
\caption{Comparison of AUC convex hulls for utility-privacy trade-off curves.}
\label{tab:4}
\end{table}

\subsection{Complexity Analysis}

As you have seen in the previous sections, the proposed scheme for image datasets performs better than the earlier works. In the case of categorical datasets, it is almost close to the performance of the best available algorithm. In addition to performance, the proposed algorithm has advantages over other algorithms in terms of complexity and convergence time. The absence of a GAN structure in the proposed scheme prevents convergence and stability problems \cite{goodfellow2016nips,kodali2017convergence,barnett2018convergence}, which makes convergence faster than GAN-based algorithms. For the categorical dataset, although our performance is slightly worse than the performance of the algorithm in \cite{mandal2022uncertainty}, as shown in Table \ref{tab:2}, our obfuscator parameters are about $25\%$ of the parameters in the proposed network in \cite{mandal2022uncertainty}. Therefore, our network structure is simpler and converges faster. Since \cite{mandal2022uncertainty} has used the GAN structure, it should also train the adversary and the utility provider in addition to the obfuscator, which increases the number of parameters.

\begin{table}
\centering
\resizebox{\columnwidth}{!}{
\begin{tabular}{l|c|c}
\hline 
\rule[-1ex]{0pt}{2.5ex} Method & Components & Number of Parameters\\ 
\hline 
\rule[-1ex]{0pt}{2.5ex} Mandal et al. \cite{mandal2022uncertainty} & Obfuscator, Utility Provider, and Adversary & $338,214$ \\ 
\hline 
\rule[-1ex]{0pt}{2.5ex} Proposed Scheme & Obfuscator & $88,494$ \\  
\hline 
\end{tabular} 
}
\caption{Comparison of the complexity of the proposed scheme with the method presented in \cite{mandal2022uncertainty} in terms of the number of parameters.}
\label{tab:2}
\end{table}

\section{Conclusion}
\label{sec:conclusion}
This paper introduced a private method of data publishing using a structure based on an autoencoder. The simulation results show that the proposed method establishes a good balance in trade-off between utility and privacy compared to previous methods. Moreover, the presented method has two main advantages over the earlier methods. First, each data provider can adjust the privacy and utility level; secondly, there is no need to specify the private feature for the obfuscation design.

\section*{Acknowledgment}

This work was partially supported by Iran National Science Foundation (INSF) under contract No. 97011231.

\end{document}